\documentclass{article}

\usepackage[preprint]{neurips_data_2023}
\usepackage[utf8]{inputenc} % allow utf-8 input
\usepackage[T1]{fontenc}    % use 8-bit T1 fonts
\usepackage{hyperref}       % hyperlinks
\usepackage{url}            % simple URL typesetting
\usepackage{booktabs}       % professional-quality tables
\usepackage{amsfonts}       % blackboard math symbols
\usepackage{nicefrac}       % compact symbols for 1/2, etc.
\usepackage{microtype}      % microtypography
\usepackage{xcolor}         % colors
\usepackage{graphicx}
\usepackage{floatrow}

\newfloatcommand{capbtabbox}{table}[][\FBwidth]

\title{aiMotive Dataset: A Multimodal Dataset for Robust Autonomous Driving with Long-Range Perception}

\author{
    Tam\'{a}s Matuszka, Iv\'{a}n Barton, Ádám Butykai, P\'{e}ter Hajas, Dávid Kiss \\
    \textbf{Domonkos Kovács, S\'{a}ndor Kuns\'{a}gi-M\'{a}té, Péter Lengyel, G\'{a}bor N\'{e}meth} \\
    \textbf{Levente Pető, Dezső Ribli, D\'{a}vid Szeghy, Szabolcs Vajna, B\'{a}lint Varga} \\
    aiMotive \\
    Budapest, Hungary \\
    \texttt{\url{https://aimotive.com}} \\
}

\begin{document}

\maketitle

\begin{abstract}
  Autonomous driving is a popular research area within the computer vision research community. Since autonomous vehicles are highly safety-critical, ensuring robustness is essential for real-world deployment. While several public multimodal datasets are accessible, they mainly comprise two sensor modalities (camera, LiDAR) which are not well-suited for adverse weather. In addition, they lack far-range annotations, making it harder to train neural networks that are the base of a highway assistant function of an autonomous vehicle. Therefore, we introduce a multimodal dataset for robust autonomous driving with long-range perception. The dataset includes 176 scenes with synchronized and calibrated LiDAR, camera, and radar sensors covering a 360-degree field of view. The collected data was captured in highway, urban, and suburban areas during daytime, night, and rain and is annotated with 3D bounding boxes with consistent identifiers across frames. Furthermore, we trained unimodal and multimodal baseline models for 3D object detection. Data and code are available at \url{https://github.com/aimotive/aimotive_dataset}.
\end{abstract}

%------------------------------------------------------------------------
\section{Introduction}
\label{intro}

A large number of datasets for 3D object detection applied in autonomous driving have been released in the last few years \citep{kitti, argo, huang2018apolloscape, pham20203d, patil2019h3d, nuscenes}. Most datasets have the common property of including sensor data from different modalities, such as cameras and LiDAR. In this way, a 360-degree field-of-view (FOV) can be covered around the ego vehicle. 3D object detection datasets can be split into groups based on the coverage around the ego vehicle and sensor redundancy. While numerous datasets are publicly available, they either do not provide sensor redundancy (i.e. sensor coverage by at least two sensor modalities) which is essential for robust autonomous driving or rely only on camera and LiDAR sensors that are not perfectly applicable in adverse weather (see Table~\ref{tab:datasets} for the properties of several popular datasets grouped based on sensor coverage and redundancy). This issue can be solved by utilizing radars that is a cost-effective sensor and is not affected by adverse environmental conditions (e.g. rain or fog). Furthermore, the annotation range does not exceed 80 meters (with a few exceptions), which is insufficient for training long-range perception systems. The limitation of the annotation range can be explained by the fact that autonomous driving datasets mainly focus on urban environments while ensuring the ability to detect objects in distant regions is critical for highway assistants and therefore for autonomous driving.

Therefore, we release a multimodal dataset for robust autonomous driving with long-range perception to overcome the abovementioned limitations. The collected dataset includes 176 scenes with synchronized and calibrated LiDAR, camera, and radar sensors covering a 360-degree field of view. The data was captured in diverse geographical areas (highway, urban, and suburban) and different time and weather conditions (daytime, night, rain). We provide 3D bounding boxes with consistent identifiers across frames that enable the utilization of our dataset for 3D object detection and multiple object tracking and prediction tasks. The proposed dataset is published under CC BY-NC-SA 4.0 license, allowing the research community to use the gathered data for non-commercial research purposes. Our main contributions are the followings:
\begin{itemize}
  \item We released a multimodal autonomous driving dataset with redundant sensor coverage (including radars) and 360$^{\circ}$ FOV.
  \item Our dataset has an extended annotation range compared to existing datasets allowing the development of long-range perception systems.
  \item We trained and benchmarked unimodal and multimodal baseline models.
\end{itemize}

By releasing our dataset and models to the public, we seek to facilitate research in multimodal sensor fusion and robust long-range perception systems.

\begin{table}[t]
  \caption{Comparison of relevant datasets. Middle group: datasets with redundant 360$^{\circ}$ sensor coverage, right group: datasets with 360$^{\circ}$ view without sensor redundancy. Range refers to the perception limit of the front and back region in the case of the middle group and the front area for the right group (ego vehicle is the origin). }
  \centering  
  \begin{tabular}{c}       
    \includegraphics[width=0.97\linewidth]{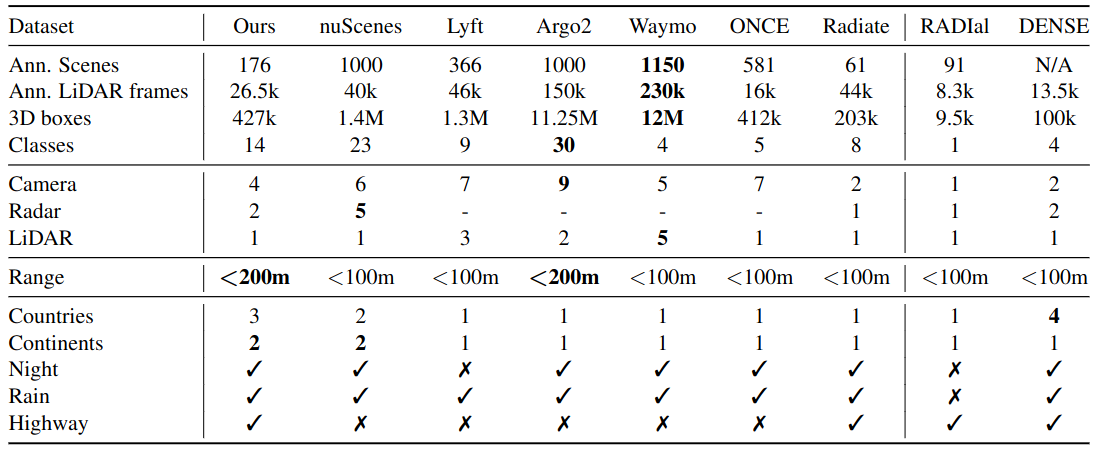}
  \end{tabular}  
  \label{tab:datasets}
\end{table}

%------------------------------------------------------------------------
\section{Related work}
\label{gen_inst}

One of the most influential datasets is KITTI by \citet{kitti}, which generated interest in 3D object detection in autonomous driving. The KITTI dataset contains 22 scenes recorded in Karlsruhe, Germany. The sensor setup consists of front cameras and a roof-mounted LiDAR. The perception range of the released dataset is less than 100 meters, and no 360-degree FOV is provided. In addition, the footage was recorded only in the daytime.

Several popular 3D object detection datasets provide a 360$^{\circ}$ FOV with sensor redundancy. Among these, nuScenes \citep{nuscenes} is the most similar dataset to our work, including full sensor redundancy for the entire sensor setup. However, a 32-beam LiDAR with a relatively sparse point cloud and limited perception range was used during the recording process, resulting in a shorter perception limit than 100 meters (i.e. there are no annotated objects with a distance larger than 100 meters from the ego vehicle at the moment when the given frame was annotated). The sensor data has been recorded in urban environments (Boston, USA, Singapore) and lacks footage on highways. Waymo Open Dataset\citep{waymo} is the first large-scale autonomous driving 3D object detection data collection with 360$^{\circ}$ FOV, including more than 1000 scenes and 12M annotated objects. The main shortcoming of this dataset is the limited perception range and sensor suite. The recently released Argoverse2 Sensor \citep{wilson2021argoverse} dataset utilized the experiences gained from hosting several challenges using the Argoverse \citep{argo} dataset. Argoverse2 has a similar scale as Waymo Open Dataset but with an extended annotation range. The disadvantage of the dataset compared with our solution is the lack of radar sensor usage and the diversity of recording locations (see Table~\ref{tab:datasets}). Both Lyft Level 5 perception dataset \citep{lyft} and ONCE \citep{mao2021one} have recordings from only one country, without using any radars, and do not contain annotated objects in distant areas. Radiate \citep{radiate} uses three different sensor modalities and contains a large amount of annotated keyframes in adverse weather (e.g. fog, rain, snow). The paper's main contribution is the release of a high-resolution radar dataset. However, the perception range is limited (i.e. less than 100 m), and other sensor modalities are constrained (32-beam LiDAR with very sparse point cloud, only front camera with low-resolution images).

Another group of datasets also provides 360-degree coverage without ensuring sensor redundancy which is essential for robust autonomous driving. RADIal\citep{radial}, similar to Radiate, employs a high-definition radar for sensing in 360$^{\circ}$. The recorded data covers a wide range of geographical areas; however, the sensor setup is restricted to only three sensors. Furthermore, only a limited amount of annotated objects (less than 10k) are contained in the dataset. DENSE \citep{dense} also focuses on data collected in severe weather. The paper describes a unique sensor setup consisting of a thermal camera, gated cameras, and a spinning LiDAR. Even though a diverse set of sensors is mounted to the recording car, sensor redundancy is not ensured in the case of the dataset. Moreover, the annotated area is limited due to the challenging weather conditions.
 
Our dataset has an advantage over the existing related work, as is seen in Table~\ref{tab:datasets}. The proposed dataset combines sensor redundancy with a long perception range in diverse environments, which is not assured by previously published 3D object detection datasets. Ensuring these properties are required for training neural networks that can serve as a base for robust autonomous driving software operating in different environments.

\begin{figure}
\begin{floatrow}
\ffigbox{%
  \includegraphics[width=1.13\linewidth]{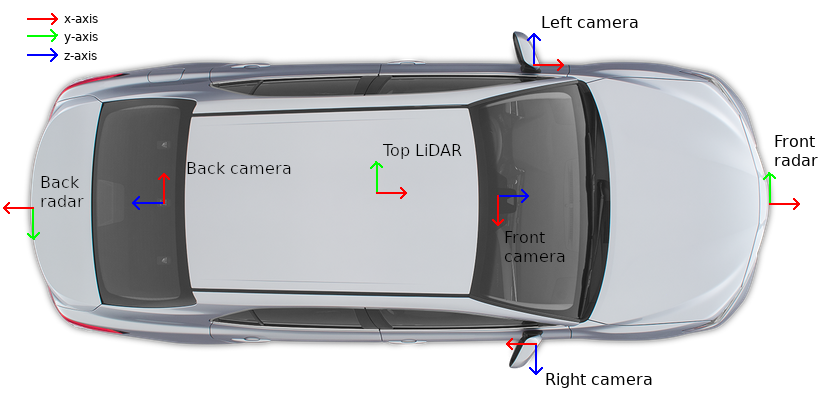}
  
}
{%
  \caption{Sensor setup and coordinate systems.}%
  \label{fig:sensor_setup}
}

\capbtabbox{%
  \begin{tabular}{c} 
    \includegraphics[width=0.823\linewidth]{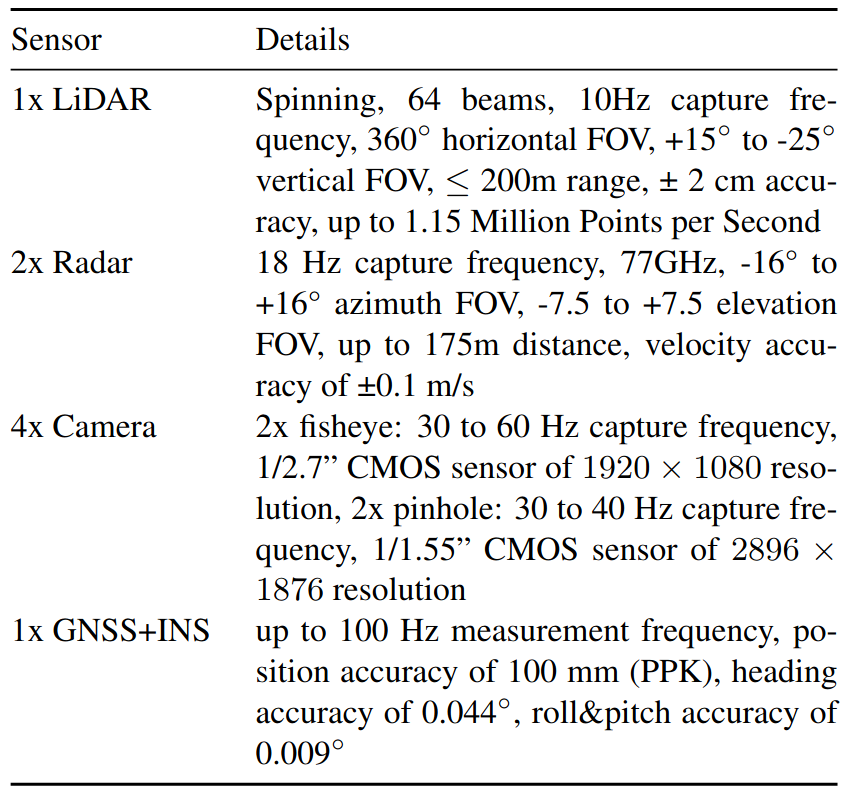} \\
  \end{tabular}
}{%
  \caption{Description of used sensors.}%
  \label{tab:sensor_desc}
}
\end{floatrow}
\end{figure}

%------------------------------------------------------------------------
\section{aiMotive Multimodal Dataset}
\label{sec:dataset}

Our multimodal dataset comprises 15s long scenes with synchronized and calibrated sensors. The dataset provides a 360$^{\circ}$ FOV with the help of a redundant sensor layout. Thus, the surrounding area of the ego vehicle is recorded by at least two different sensors. Since the annotated 3D bounding boxes have consistent identifiers across frames within a scene, the dataset can be used for 3D object detection and multiple object tracking and prediction tasks. In addition, a considerable amount of annotations (about 25\%) are located in the far-distance region ($\geq75m$) concerning the ego vehicle. Due to this property and the redundant sensor setup, our dataset can facilitate research in multimodal sensor fusion and robust long-range perception systems.

\begin{figure}
\begin{floatrow}
\capbtabbox{%
  \begin{tabular}{c} 
    \includegraphics[width=0.9\linewidth]{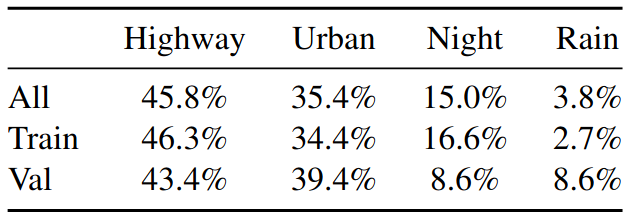}
  \end{tabular}
}
{%
  \caption{Data distribution w.r.t weather and environment.}%
  \label{tab:data_distr}
}
\capbtabbox{%
  \begin{tabular}{c} 
    \includegraphics[width=0.823\linewidth]{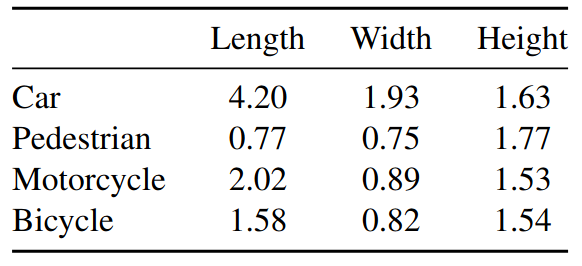} \\
  \end{tabular}
}{%
  \caption{Average cuboid dimensions (m).}%
  \label{tab:avg_cuboid}
}
\end{floatrow}
\end{figure}

%------------------------------------------------------------------------
\subsection{Data collection}
\label{datacoll}
The data was collected in three countries on two continents with four cars to provide a diverse dataset. The recordings have taken place in California, US; Austria; and Hungary using three Toyota Camry and one Toyota Prius. The recording phase of the footage was spread across a year to gather data in different seasons and weather conditions. As a result, our dataset consists of a diverse set of locations (highway, suburban, urban), times (daytime, night), and weather conditions (sun, cloud, rain, glare). The data collection method has satisfied the requirements given by the Institutional Review Board approval.

\subsection{Sensor setup}
\paragraph{Sensor layout.} The data was recorded using a roof-mounted, rotating 64-beam LiDAR, four cameras, and two long-range radars, providing 360$^{\circ}$ coverage with sensor redundancy. The localization was based on a high-precision GNSS+INS sensor. Additional details can be found in Figure~\ref{fig:sensor_setup} and Table~\ref{tab:sensor_desc}.

\paragraph{Synchronization.} All of the recorded sensor data are synchronized. The LiDAR and radars share the same timestamp source. Our cameras capture images using the rolling shutter method, which scans the environment rapidly instead of capturing the image as a snapshot of the entire scene at a single time moment. Since the used cameras capture the scene row by row, the camera timestamp is approximately the exposure time when the middle row is captured.

\paragraph{Coordinate systems.} The dataset uses five coordinate systems. Namely global, body, radar, camera, and image coordinate systems. We have used ECEF \citep{snay1999modern} as the global coordinate system and provided a 6-DOF ego-vehicle pose for each annotated frame. The reference coordinate system used for defining the annotated objects is called the body coordinate system and is assigned to the vehicle body. The origin is the projected ground plane point under the center of the vehicle’s rear axis at nominal vehicle body height and zero velocity. The radar coordinate system uses the same axes as the body coordinate system ($x$-axis positive forward, $y$-axis positive to the left, and $z$-axis positive upwards). The LiDAR point cloud was transformed into the body coordinate system as a preprocessing step. The origin of the camera coordinate system is the camera’s viewpoint, and the axes are defined the same as the OpenCV\citep{opencv} camera coordinate system ($x$-axis positive to the right, $y$-axis positive downwards, $z$-axis positive forward). Camera-to-body and radar-to-body transformations can be performed using camera and radar extrinsic matrices. We utilized OpenCV's image coordinate system to render annotations using intrinsic matrices to project from camera coordinates to image coordinates.

%------------------------------------------------------------------------
\subsection{Ground truth generation}
\label{gt}
We used two approaches for generating ground truth labels: an automatic annotation method for training data generation and manual annotation for creating validation data. The automatic annotation relies on LiDAR measurements and searches possible candidates in the entire point cloud of a 15s long sequence. Non-causal object tracking (i.e. both directions in time) including the association of new detections to the existing tracks is realized in the 2.5D descriptor space with the joint probability distribution of the modeled detection uncertainty and object dynamics. Utilizing the informative point cloud along with the physical constraints, the consecutive detections (i.e. positions and orientations) of the same object can be optimized recursively. In this way, the point cloud of a given object can be accumulated from different views. As the tracked object's trajectory becomes more accurate with the optimization steps, the model point cloud of the detected object becomes sharp; thus, a bounding box can be fitted on it. The annotated sequences were manually quality-checked based on multiple criteria. This inspection checks the position (<10\% divergence between the object centroid and the cuboid centroid), orientation (within 5 degrees of precision), and size (<10\% divergence between the object size and the cuboid size) of the amodal bounding boxes projected back to all available cameras. Since the sensors are mounted in different positions, some annotated objects may be occluded on specific sensors. The manual quality checking is performed on the scene level. Some label noise still might be in the dataset even though we aimed to minimize it via human validation. In this way, we selected sufficiently accurately labeled recordings, and most scenes with erroneous annotations were discarded. 

In the case of the validation set, we hired manual annotators to label objects on the recorded sensor data. The human annotators used LiDAR and camera sensor data during the annotation phase to fit cuboids on any object of interest appearing on the camera images. For the cuboid sizes, annotators used default dimensions. If the default dimensions do not match the size of a given object in the point cloud or on the images, the annotators refined the non-matching dimensions of the given cuboid based on their own decision. The manual labor also ensured that the cuboid axes aligned with the object orientation within 5 degrees of precision.

The manually or automatically annotated objects belonging to 14 classes are represented as 3D cuboids with some additional physical properties. Each labeled bounding box has a 3D center point, 3D extent (length along the horizontal $x$-axis, width along the vertical $y$-axis, height along the $z$-axis), orientation (represented as a quaternion), relative velocity, and a unique track ID. Furthermore, we provide 2D bounding boxes utilizing a pretrained FCOS \citep{tian2019fcos} detector. The 2D-3D annotations are associated using the Hungarian algorithm \citep{kuhn1955hungarian} for allowing the utilization of 2D-3D consistency or semi-pseudo-labeling \citep{matuszka2022novel}. The resulting dataset was anonymized using Dashcam-Cleaner\footnote{https://github.com/tfaehse/DashcamCleaner}.

%------------------------------------------------------------------------
\subsection{Dataset analysis}
The dataset includes 26 583 annotated frames with sensor data from multiple modalities, split into 21 402 train and 5 181 validation frames (151/25 train/val scenes). The scenes were recorded in diverse weather and environmental conditions. See Table~\ref{tab:data_distr} for the data distribution. Since the training and validation splits were generated with different methods, some distribution shifts between the data partitions might arise. We investigated the distribution of scenes concerning weather/environment and object count/dimensions and found that training and validation splits are similar in proportion.

\begin{figure}[t]
\begin{floatrow}
\ffigbox{%
  \includegraphics[width=1.0\linewidth]{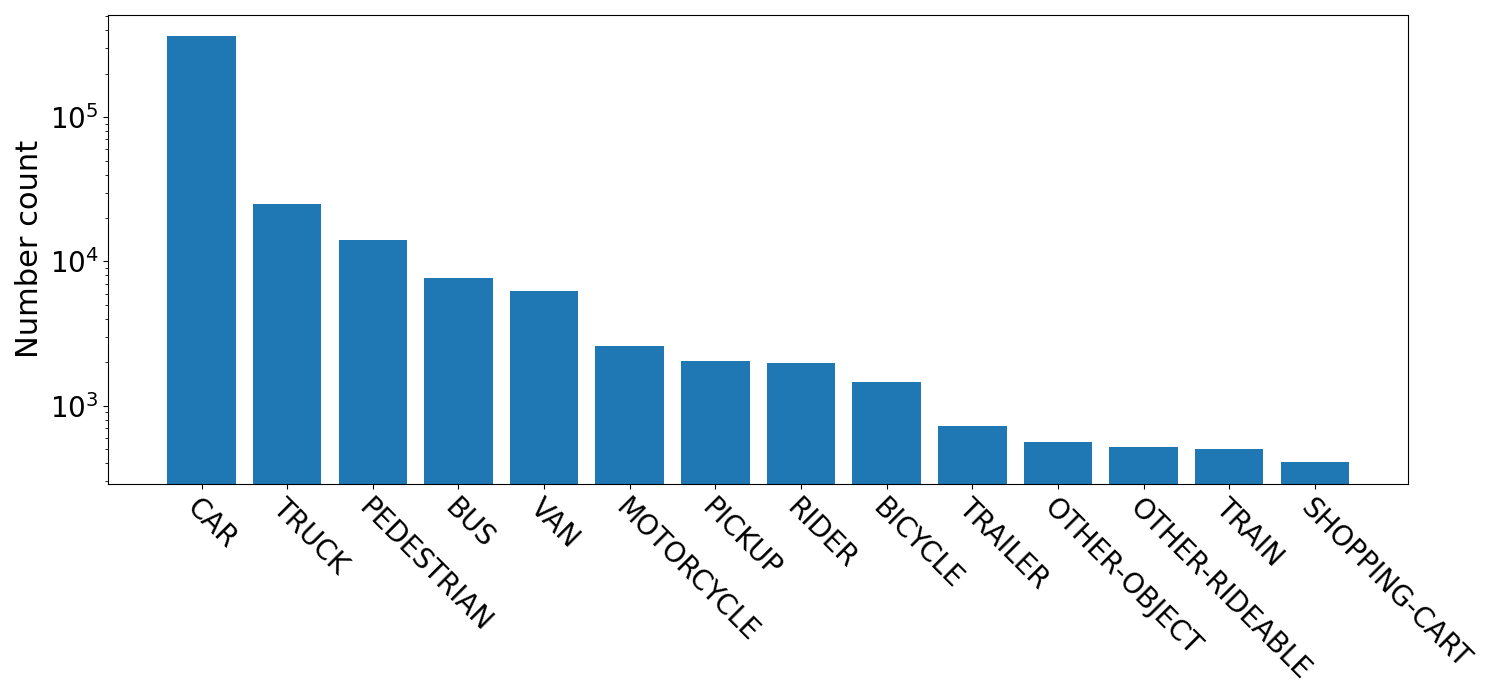}
  
}
{%
  \caption{Class distribution.}%
  \label{fig:class_dist}
}
\ffigbox{%
    \includegraphics[width=1.0\linewidth]{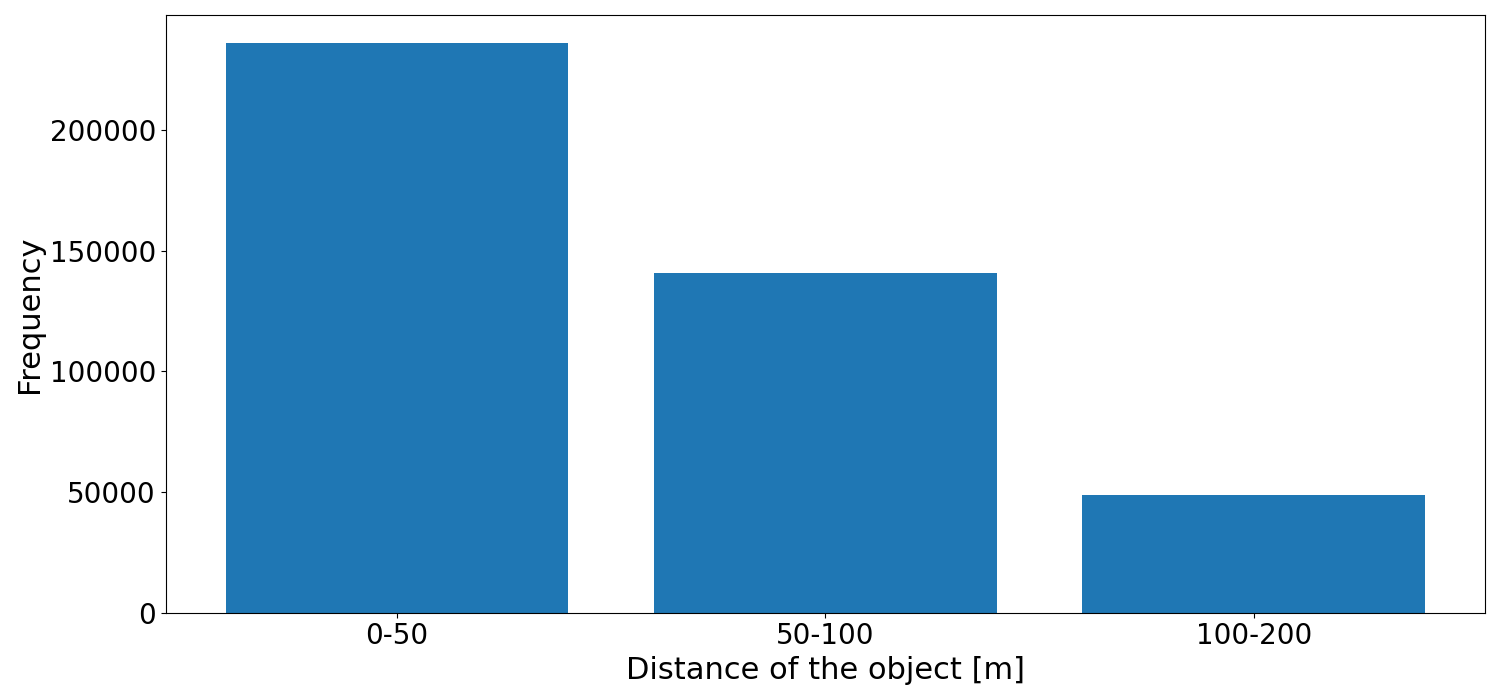} \\
}{%
  \caption{Distance distribution of annotated objects.}%
  \label{fig:dist}
}
\end{floatrow}
\end{figure}

The dataset contains more than 425k objects organized into 14 categories. See the class distribution in Figure~\ref{fig:class_dist}. The distance distribution of the annotated objects is visualized in Figure~\ref{fig:dist}. About 24\% of the cuboids are beyond 75 m, Argoverse2 has about 14\%, Waymo, nuScenes, and ONCE have less than 1\%. This property enables the training of long-range perception systems with the help of our dataset. 

\begin{figure}[t]
\begin{floatrow}
\ffigbox{%
  \includegraphics[width=0.9\linewidth]{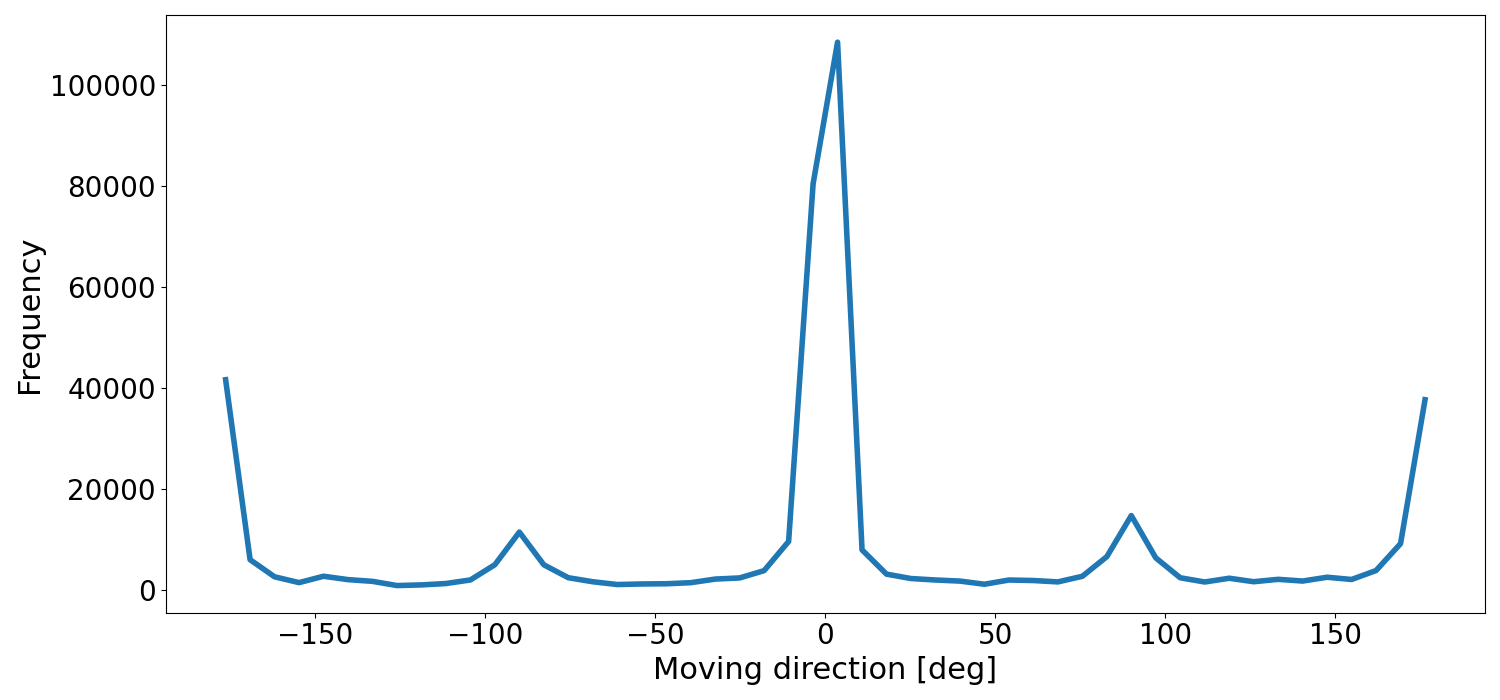}
}
{%
  \caption{Orientation distribution of annotated objects.}%
  \label{fig:orientation}
}
\capbtabbox{%
  \begin{tabular}{c} 
    \includegraphics[width=0.9\linewidth]{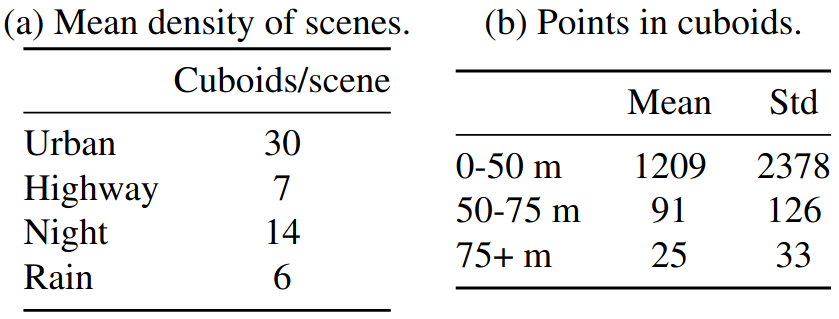} \\
  \end{tabular}
}{%
  \caption{Additional statistics of the annotations.}%
  \label{tab:add_stats}
}
\end{floatrow}
\end{figure}

Several additional statistics of the generated dataset are described by Table~\ref{tab:avg_cuboid}, Table~\ref{tab:add_stats}, and Figure~\ref{fig:orientation}. The average cuboid dimensions for distinguished classes help to understand how precise the cuboids are per class. The number of average cuboids per environment indicates how crowded the scenes are. The percentage of empty boxes beyond 50 m and 75 m after the annotation process is 4.2\% and 5.4\%, respectively, as opposed to the conventional benchmarks where almost 50\% of objects beyond 50 m contain zero LiDAR points \citep{gupta2023far3det}. 

%------------------------------------------------------------------------
\section{Experiments}
\label{sec:experiments}
We trained several 3D object detection baselines on our dataset utilizing publicly available models. In order to exploit annotations located in distant areas, we defined the target grid as $[-204.8,204.8]$ m in longitudinal and $[-25.6,25.6]$ m in lateral directions. We mapped the 14 classes included in the dataset into four categories (car, truck/bus, motorcycle, pedestrian), then evaluated the model performance using the all-point and 11-point interpolated Average Precision (AP) metrics \citep{everingham2010pascal} in Bird's-Eye-View (BEV) space in a class agnostic manner. The Hungarian method \citep{kuhn1955hungarian} is used for associating ground truth and predictions with a 0.3 IoU threshold. We selected a small IoU value for the association threshold to handle displacement errors which are especially frequent in distant regions in BEV. Furthermore, the Average Orientation Similarity (AOS) \citep{geiger2012we} metric is utilized for evaluating the performance of the models in terms of orientation prediction. 

\begin{table}[t]
  \centering  
  \begin{tabular}{c}       
    \includegraphics[width=0.99\linewidth]{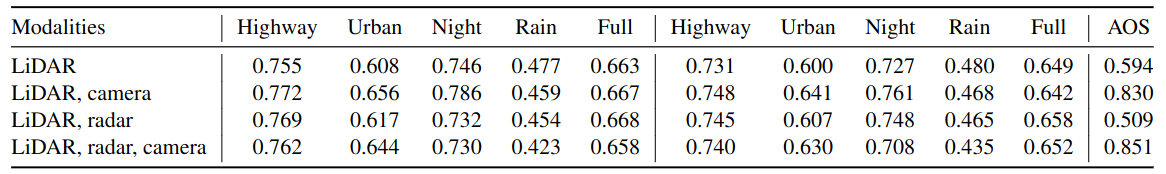}
  \end{tabular}
  \caption{Comparison of baseline models. First group: all-point AP metric, second group: 11-point interpolation AP metric, third group: AOS metric averaged over val set. }
  \label{tab:metrics}
\end{table}

\begin{figure}[t]
  \centering
   \includegraphics[width=1.0\linewidth]{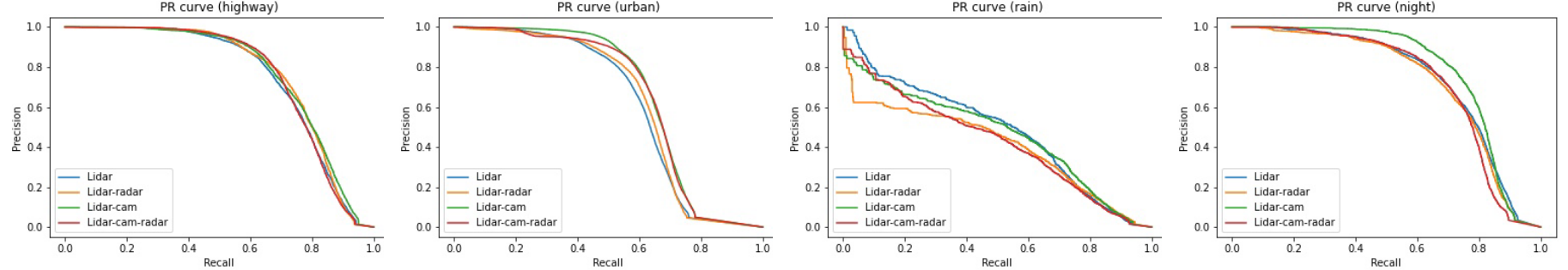}
   \caption{PR curves of baseline models.}
   \label{fig:pr-curves}
\end{figure}

\subsection{Baseline models}
Our baseline models are based on VoxelNet \citep{zhou2018voxelnet}, BEVDepth \citep{li2022bevdepth}, and BEVFusion \citep{liu2022bevfusion} for LiDAR, camera, and multimodal models. Since BEVFusion does not use radar sensors, we designed a simple solution for LiDAR-radar fusion. Namely, we treated the radar point cloud as a regular LiDAR point cloud. After a point cloud merging step, data from different modalities can be processed by VoxelNet as if it was a regular LiDAR point cloud.

VoxelNet can operate on the point cloud directly and consists of three main parts. The Voxel Feature Encoder (VFE) is responsible for encoding raw point clouds at the individual voxel level. VoxelNet utilizes stacked VFE layers, and their output is further processed by a middle convolutional neural network (CNN) to aggregate voxel-wise features. The final component performing the 3D object detection is the region proposal network \citep{ren2015faster}.

BEVDepth is a camera-only 3D object detection network that provides reliable depth estimation. The main observation of the authors is that recent camera-only 3D object detection solutions utilizing pixel-wise depth estimation generate suboptimal results due to inadequate depth estimation. Therefore, explicit depth supervision encoding intrinsic and extrinsic parameters is utilized. In addition, a depth correction subnetwork is introduced using sparse depth data from a LiDAR point cloud to provide supervision for the depth estimation network.

The main contribution of BEVFusion is the utilization of the BEV space as the unified representation for camera and LiDAR sensor fusion. The image backbone proposed by BEVFusion explicitly predicts a discrete depth distribution for each image pixel, similar to BEVDepth (without the depth correction subnetwork). Then, a BEV pooling operator is applied on the 3D feature point cloud that is later flattened along the $z$-axis to get a feature map in BEV. The point cloud produced by a LiDAR is processed the same way as in the case of VoxelNet. Then, the two BEV feature maps are fused by a CNN. Finally, the detection heads are attached to the output of the fusion subnetwork.

\begin{table}[t]
  \centering  
  \begin{tabular}{c}       
    \includegraphics[width=0.99\linewidth]{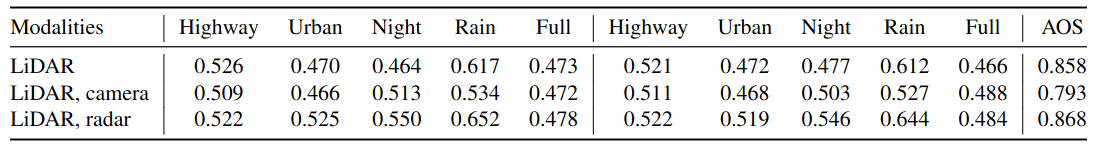}
  \end{tabular}
  \caption{Comparison of baseline models in the distant region (\textgreater75m). First group: all-point AP metric, second group: 11-point interpolation AP metric, third group: AOS metric averaged over the validation set.}
  \label{tab:metrics-distant}
\end{table}

\begin{figure}[t]
  \centering
   \includegraphics[width=1.0\linewidth]{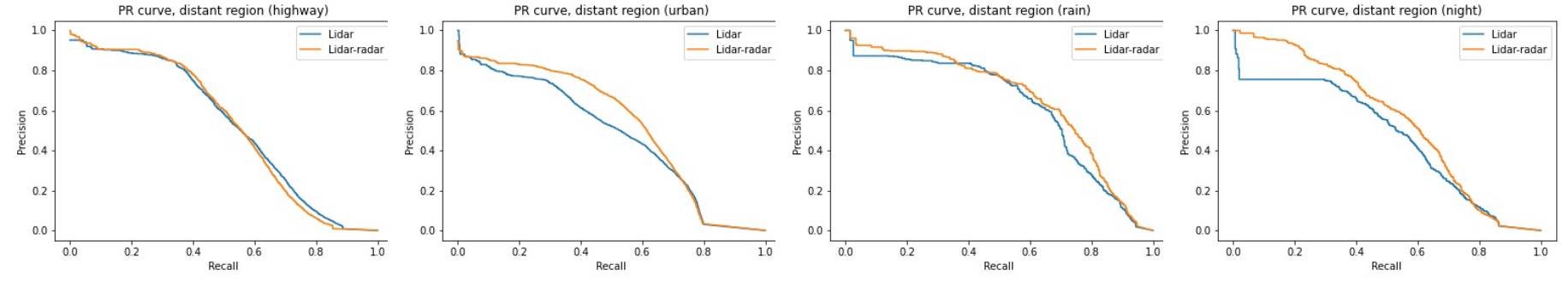}
   \caption{PR curves of baseline models in distant region (\textgreater75m).}
   \label{fig:pr-curves-distant}
\end{figure}

\subsection{Implementation details}
The LiDAR components of the baseline models use HardSimpleVFE \citep{yan2018second} as the Voxel Feature Encoder and SparseEncoder \citep{yan2018second} as the middle encoder CNN. The image components adopt Lift-Splat-Shoot \citep{philion2020lift} as an image encoder with a ResNet-50 backbone followed by a Feature Pyramid Network \citep{lin2017feature} for leveraging multi-scale features. An additional depth correction network is also part of the image stream, inspired by BEVDepth. In the case of multimodal models, features from different modalities are fused using a simple fusion subnetwork consisting of convolution and Squeeze-and-Excitation \citep{hu2018squeeze} blocks. Finally, a CenterPoint \citep{yin2021center} head is responsible for detecting objects from the BEV features, both in unimodal and multimodal cases.

Since our goal is not to develop state-of-the-art models in this work but to facilitate multimodal object detection research, we used the hyperparameters provided by BEVDepth \footnote{https://github.com/Megvii-BaseDetection/BEVDepth} without any heavy parameter tuning. We adapted the grid resolution to enable long-range detection, then trained the models for 16k iterations (3 epochs) using batch size 4 with a learning rate of $6.25e^{-5}$ using flip, rotation, and scale augmentations in the BEV feature space. We used an NVIDIA A100 TensorCore GPU for neural network training. The models are implemented using mmdetection3d\footnote{https://github.com/open-mmlab/mmdetection3d} and are publicly available on the dataset repository.

\subsection{Experimental results}
\label{subsec:exp_results}
The performance comparison of the baseline models on different metrics is described in Table~\ref{tab:metrics}. Since the literature describes several examples \citep{qian20223d, liu2022bevfusion} of the superiority of LiDAR-only unimodal solutions over camera-only models, we did not train a camera-only baseline. As the table describes, every multimodal model overperforms the LiDAR-only baseline in highway and urban environments in non-adverse weather and time. The additional sensor signals significantly increase detection performance in the dense urban environment. However, the unimodal baseline performs best in heavy rain, where one would think a radar signal should help to increase performance. This phenomenon suggests that more sophisticated radar fusion techniques might be beneficial for enhancing multimodal models.

Cameras play a crucial role in terms of orientation prediction. The models without RGB images struggle to consistently keep the orientation, especially in the case of large vehicles. This flickering effect is less visible for models using camera sensors. The model using all modalities performs best on the AOS metric.

\begin{table}[t]
  \centering  
  \begin{tabular}{c}       
    \includegraphics[width=0.65\linewidth]{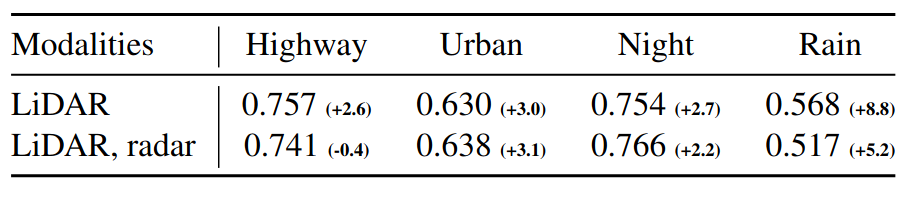}
  \end{tabular}
  \caption{Effects of longer training.}
  \label{tab:metrics-long}
\end{table}

Surprisingly, the model using LiDAR + camera modalities overperforms all other models in the night and urban environments by a large margin. We investigated the learning curves and found that increasing the number of training steps can help to enhance performance further. To validate our hypothesis, we trained our models for five additional epochs. Unfortunately, models using camera sensors became unstable after the third epoch and caused an explosion in the depth loss. Table~\ref{tab:metrics-long} describes the result of the longer training process using the 11-point interpolation AP metric. A solid improvement can be seen in all environments, especially on the rainy validation set ($+8.8$/$+5.2$ AP for LiDAR and LiDAR+radar models, respectively). This can be explained by the fact that the detection heatmaps became sharper after more prolonged training. Blurry heatmaps were responsible for lower AP metrics in the case of the first group of baseline models. The blurring effect on the heatmap was distinctly visible around the ego car in heavy rain due to LiDAR reflections from raindrops.

In order to validate the long-range perception capabilities of the baseline models, we benchmarked the longer-trained models on distant object detection. Detections and ground truth were filtered out where the distance from the ego car was less than 75 meters. The results are summarized in Table~\ref{tab:metrics-distant}. All models except LiDAR + camera perform similarly in the highway environment without any significant difference in performance. However, the model with additional radar signals significantly overperforms the LiDAR-only baseline in all other domains. The fact that radar sensors provide reliable and accurate signals for perceiving objects in distant areas, even in adverse weather, can be leveraged for boosting 3D object detector performance, as can be seen in Figure~\ref{fig:pr-curves-distant}. A similar effect can be observed in a dense urban environment where radar signals are utilized by the multimodal baseline that resulted in a significant performance increase in long-range perception ($+5.5$/$+4.7$ all-point / 11-point interpolation AP). The overall performance on the whole dataset of the multimodal models is also better considering the 11-point interpolation metric. 

The training results demonstrate that our dataset can serve as a base for multimodal long-range perception neural network training. Advanced evaluation techniques such as test-time augmentation or model ensembling could lead to further improvements. However, none of them were applied during the evaluation method. Table~\ref{tab:metrics-long} suggests that further improvements in sensor fusion methods are needed for fully leveraging each modality, and our naive approach provides a suboptimal solution (especially in the case of heavy rain). Nevertheless, we hope the research community will find our dataset valuable and can build on our baselines, and will significantly improve its performance.

%------------------------------------------------------------------------
\section{Additional proposed tasks}
We propose additional tasks benefiting from our dataset besides 3D object detection. Since unique track IDs are provided, end-to-end long-range multiple object tracking models can also be trained with the help of the dataset. Multiple Object Tracking Accuracy (MOTA) and Multiple Object Tracking Precision (MOTP) metrics can be used for evaluating model performance.

Another proposed task is motion prediction. The ego-motion is included in the dataset and the trajectories of exo-objects can be computed using the unique track IDs. We propose a specific case of motion prediction, namely lead car prediction, which is essential for autonomous driving functions such as Automatic Emergency Braking or Adaptive Cruise Control. The lead cars can be determined by the intersection of ego-motion and the trajectories of exo-objects. The proposed task is to detect and predict the current and future lead vehicles. The model performance can be measured using precision and recall metrics.

The dataset also includes high-quality GNSS-INS sensory data, thus enabling the training and benchmarking of various odometry algorithms. Finally, the dataset can be used for contrastive representation learning. A similar representation can be learned for different sensor modalities corresponding to the same frame in a self-supervised manner which might be a valuable method for a model to acquire a consistent world representation from multiple sensors.

%------------------------------------------------------------------------
\section{Conclusion}
\label{sec:conclusion}
In this paper, we present a multimodal dataset for robust autonomous driving with long-range perception. Our diverse dataset recorded in three countries on two continents includes sensor data from LiDAR, radars, and cameras providing redundant 360-degree sensor coverage. The dataset contains a large number of annotated objects in distant areas, allowing the development of multimodal long-range perception neural networks. In addition, we developed several unimodal and multimodal baseline models and compared their performance on the proposed dataset based on different criteria. We showed that our dataset is suitable for training multimodal long-range perception neural networks leveraging the advantages of the recorded sensor modalities. 

\paragraph{Limitations.} The sensor setup results in a synchronization limitation caused by the fact that sensors differ in the temporal recording method. Since the rotating LiDAR and rolling shutter cameras have different measurement methods, there are some discrepancies in temporally discretized annotations. This phenomenon is mostly visible when the relative speed difference between the ego and exo car is large. Furthermore, though the entire FOV is covered by at least two sensor modalities, the dataset lacks side radars which would be beneficial for providing full sensor coverage. Fisheye cameras used on the sides give distorted images and shorter visibility range than pinhole cameras which might limit the performance for long-range detection in the side areas.

\newpage

\paragraph{Future work.} We aim to extend our collected dataset with additional environmental and weather conditions. Furthermore, we will conduct more in-depth experiments regarding sensor fusion for multimodal neural networks to overcome the detected weaknesses of the proposed fusion method. We seek to facilitate research in multimodal sensor fusion and robust long-range perception systems by releasing our dataset.

\begin{figure}[!h]
  \centering
   \includegraphics[width=0.9\linewidth]{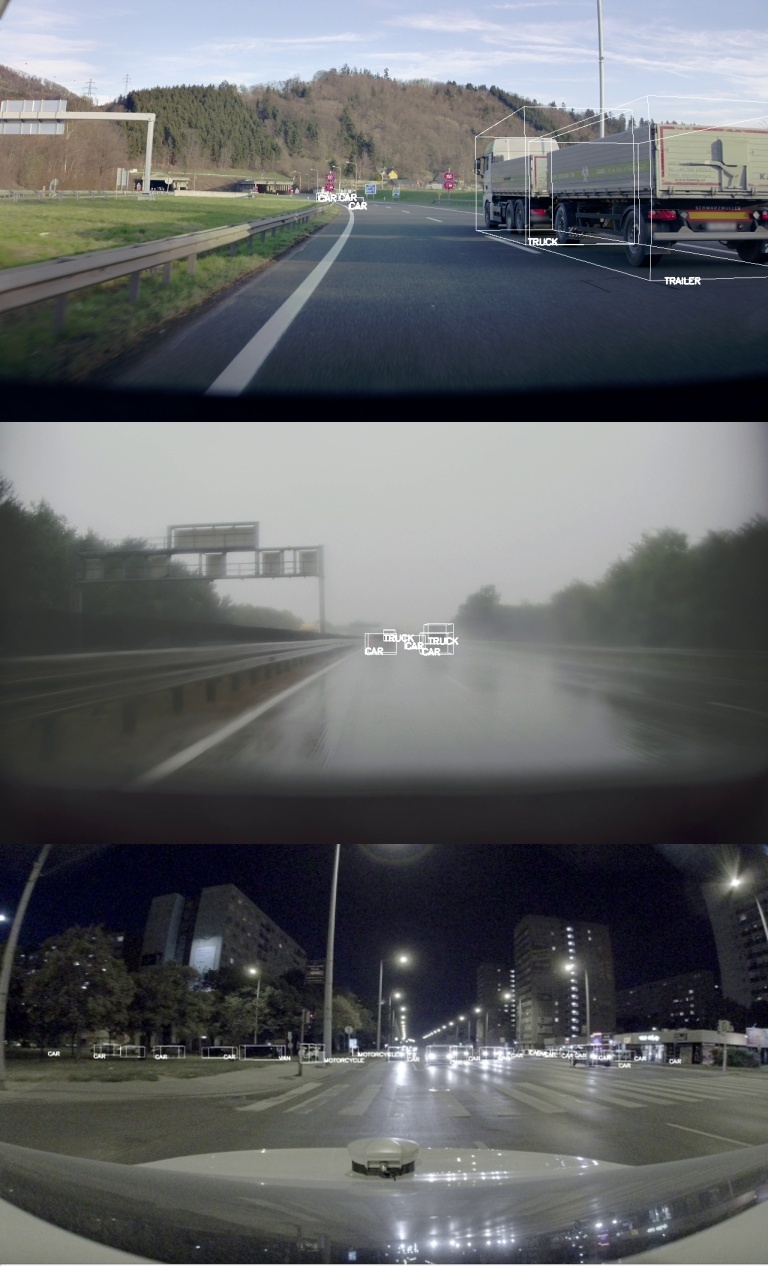}

   \caption{Example ground truth annotations from the training set (best viewed by zooming in).}
   \label{fig:annots}
\end{figure}

\bibliography{neurips_data_2023}
\bibliographystyle{neurips_data_2023}

%%%%%%%%%%%%%%%%%%%%%%%%%%%%%%%%%%%%%%%%%%%%%%%%%%%%%%%%%%%%

%\title{Supplementary Material - aiMotive Dataset: A Multimodal Dataset for Robust Autonomous Driving with Long-Range Perception}

% The \author macro works with any number of authors. There are two commands
% used to separate the names and addresses of multiple authors: \And and \AND.
%
% Using \And between authors leaves it to LaTeX to determine where to break the
% lines. Using \AND forces a line break at that point. So, if LaTeX puts 3 of 4
% authors names on the first line, and the last on the second line, try using
% \AND instead of \And before the third author name.

\newpage

\hrule height 4pt
\vskip 0.25in
\vskip -\parskip%

\begin{center}
\textbf{\LARGE Supplementary Material - aiMotive Dataset: A Multimodal Dataset for Robust Autonomous Driving with Long-Range Perception}
\end{center}

\vskip 0.29in
\vskip -\parskip
\hrule height 1pt%\p
\vskip 0.29in%

%\maketitle

%------------------------------------------------------------------------
\setcounter{section}{0}

\section{Dataset documentation (datasheets for datasets)}
\label{sec_datasheet}

\subsection{Motivation}
\paragraph{For what purpose was the dataset created?} The reason why the dataset has been created is twofold. First, while several datasets are publicly available, they either do not provide sensor redundancy (i.e. coverage by at least two sensor modalities) which is crucial for robust autonomous driving, or rely only on camera and LiDAR sensors that provide suboptimal performance in adverse weather. Second, the annotation range of these datasets does not exceed 80 meters (with a few exceptions) which is insufficient for training long-range perception systems. The limitation of the annotation range can be explained by the fact that autonomous driving datasets mainly focus on urban environments while ensuring the ability to detect objects in distant regions is critical for highway autonomous driving. Our dataset has an extended annotation range (up to 200 meters) with unique object IDs. Furthermore, it provides a 360-degree FOV using a redundant sensor layout. Due to this solution, the area around the ego vehicle is recorded by at least two different sensors. Therefore, aiMotive dataset can be used for several tasks related to long-range perception (e.g. 3D object detection, end-to-end tracking) and motion prediction.

\paragraph{Who created the dataset (e.g., which team, research group) and on behalf of which entity (e.g., company, institution, organization)?}
The dataset was created by several internal teams (Vehicle Testing \& Engineering, Calibration, aiNotate, Object Detection) on behalf of \href{https://aimotive.com}{aiMotive}.

\paragraph{Who funded the creation of the dataset?} The dataset creation was funded by aiMotive.

\subsection{Composition}
\begin{figure}[t]
  \centering
   \includegraphics[width=1.0\linewidth]{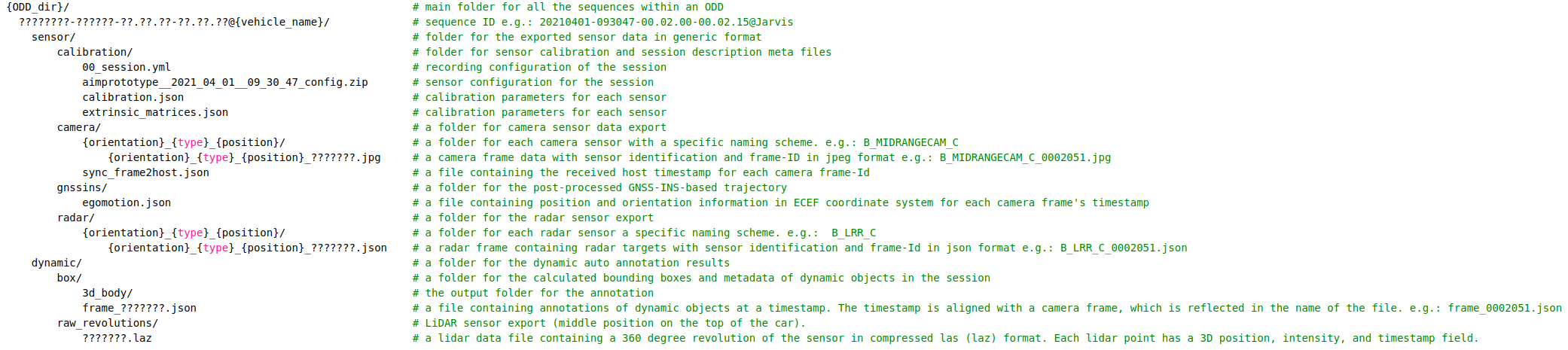}
   \caption{Structure of the dataset.}
   \label{fig:data_struct}
\end{figure}

\paragraph{What do the instances that comprise the dataset represent (e.g., documents, photos, people, countries)? Are there multiple types of instances (e.g., movies, users, and ratings; people and interactions between them; nodes and edges)?} The dataset contains multimodal sensor data (camera, radar, LiDAR, GNSS-INS), calibration, and labels. The dataset structure can be seen in Figure~\ref{fig:data_struct}. Camera sensor data is exported as JPG images for each camera. A LiDAR data file contains a 360-degree revolution of the sensor in compressed las (laz) format. Radar data, GNSS-INS data, calibration, and annotations are stored in JSON files.

\paragraph{How many instances are there in total (of each type, if appropriate)?}
The dataset includes 26 583 annotated frames with sensor data from multiple modalities. One instance consists of the following:
\begin{itemize}
  \item Four camera images (front and back cameras with pinhole camera model, left and right cameras with fisheye camera model).
  \item One LiDAR data item (360-degree revolution).
  \item Two radar sensor data item as JSON files.
  \item GNSS-INS data (ego-motion).
  \item Calibration as JSON files.
  \item Labels as JSON files.
\end{itemize}

\paragraph{Does the dataset contain all possible instances or is it a sample (not necessarily random) of instances from a larger set?} Yes, the dataset contains all possible instances.

\begin{figure}[t]
  \centering
   \includegraphics[width=0.3\linewidth]{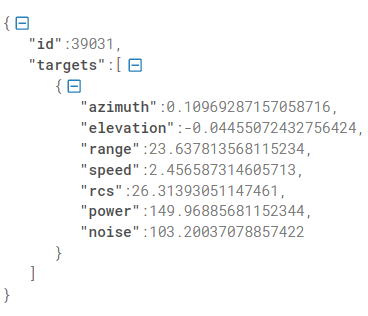}
   \caption{A sample of radar data.}
   \label{fig:radar}
\end{figure}

\paragraph{What data does each instance consist of?} The camera sensor data corresponding to one instance includes four camera images in JPG format. The back camera resolution is 1920x1216. Other cameras have 1280x704 resolution. A LiDAR data file contains a 360-degree revolution of the sensor in compressed las (laz) format. Each LiDAR point is characterized by its 3D position, intensity, and timestamp.
Radar data are stored in JSON files. The sensor measures range, radial speed, azimuth and elevation angle, and reflectivity. These values are provided by the dataset. The speed in raw radar data is the radial speed with a meter/second unit. A radar data sample can be seen in Figure~\ref{fig:radar}. The GNSS-INS data (ego-motion) can be used for transforming relative positions into an absolute coordinate system. The ego-motion-related JSON file contains the following.

\begin{itemize}
  \item key: camera frame id
  \item value is a dictionary where:
  \begin{itemize}
    \item RT\_ECEF\_body is a matrix from body coordinate system to ECEF
    \item enh\_sep is the latitude, longitude, height separation, i.e. GNSS error
    \item rph\_sep is the roll, pitch, heading separation, i.e. IMU error
    \item time: GNSS+INS device timestamp
    \item time\_host: synchronized host timestamp (from camera metadata)
  \end{itemize}
\end{itemize}

The metadata required for calibrations is stored in a JSON file. A detailed description of the utilized coordinate systems can be found in the paper. The matrices for transforming from sensor to body coordinate system (and vice versa) are stored by the data sensor key (e.g. B\_MIDRANGECAM\_C). The position of the sensor is stored by 'pos\_meter' key defined in the body coordinate system while the orientation of the sensor is described by 'yaw\_pitch\_roll\_deg' key. Camera sensors have additional metadata such as 'focal\_length\_px', 'principal\_point\_px', 'image\_resolution\_px', 'distortion\_coeffs', and 'model'.

\begin{figure}[t]
  \centering
   \includegraphics[width=0.5\linewidth]{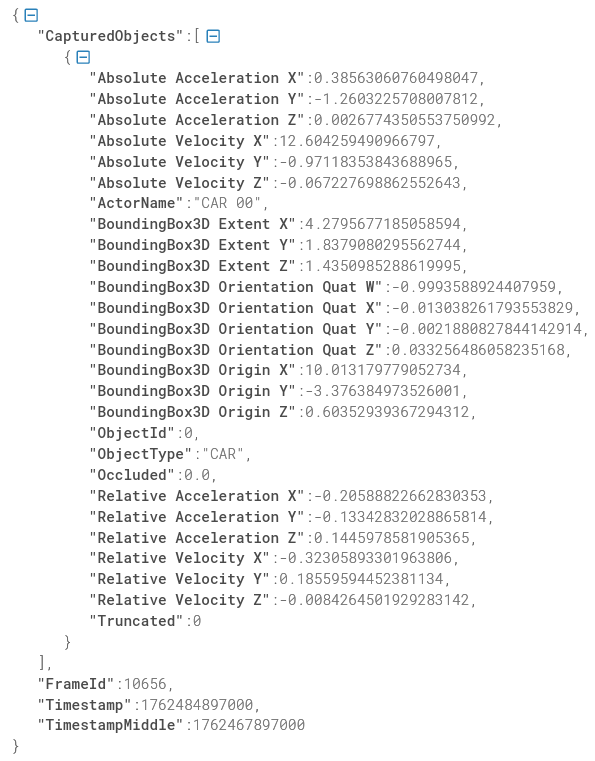}
   \caption{A sample annotation.}
   \label{fig:annot}
\end{figure}

\paragraph{Is there a label or target associated with each instance?} Yes, each data instance has an associated annotation file. The labels are stored in JSON format where an array stores the annotations of all objects corresponding to the 360-degree surroundings of the ego-vehicle. An example label can be seen in Figure~\ref{fig:annot}.

\paragraph{Is any information missing from individual instances?} Occlusion and truncation are currently not calculated for annotated objects. Work is in progress using a PC voxelization-based method. The dataset will be updated when the development is done.

\paragraph{Are relationships between individual instances made explicit (e.g., users’ movie ratings, social network links)?} There are no specific relationships between individual instances.

\paragraph{Are there recommended data splits (e.g., training, development/validation, testing)?} The dataset is split into 21 402 train and 5 181 validation frames (151/25 train/val scenes). The scenes were recorded in diverse weather and environmental conditions. See Table 3 in the paper for the data distribution. Since the training and validation splits were generated with different methods, some distribution shifts between the data partitions might arise. We investigated the distribution of scenes concerning weather/environment and object count/dimensions and found that training and validation splits are similar in proportion.

\paragraph{Are there any errors, sources of noise, or redundancies in the dataset?} The training split has been annotated with an automatic method (see description in paper). 
The annotated sequences were manually quality-checked based on multiple criteria. The following cases are marked as erroneous in the first phase of the manual quality assurance process: 
\begin{itemize}
    \item Cuboid missing from the object to be annotated (false negative). 

    \item A phantom box is applied where no object is present to be annotated (false positive). 

    \item Yaw, pitch, or roll divergence (misaligned box). 

    \item Misplaced box (position error). 

    \item More than a 10\% divergence between the object size and the cuboid size (size error). 

Multiple cuboids are applied to the same object. 
\end{itemize}

During this inspection, the amodal bounding boxes are projected back to all available cameras. The oriented bounding boxes are inspected from the top view too. The manual quality checking is performed on the scene level. When a scene has been validated by one of the manual laborers, it goes through a second phase where the manual quality-checking supervisor can either confirm it or send it back for recheck. In the case of a recheck, phase one has to be performed again. 

Some label noise still might be in the dataset even though we aimed to minimize it via human validation. In this way, we selected sufficiently accurately labeled recordings, and most scenes with erroneous annotations were discarded.

\paragraph{Is the dataset self-contained, or does it link to or otherwise rely on external resources (e.g., websites, tweets, other datasets)?} The dataset is self-contained and available on \href{https://github.com/aimotive/aimotive_dataset}{GitHub} and \href{https://www.kaggle.com/datasets/tamasmatuszka/aimotive-multimodal-dataset}{Kaggle}.

\paragraph{Does the dataset contain data that might be considered confidential (e.g., data that is protected by legal privilege or by doctor-patient confidentiality, data that includes the content of individuals’ non-public communications)?} No, the dataset does not contain confidential data.

\paragraph{Does the dataset contain data that, if viewed directly, might be offensive, insulting, threatening, or might otherwise cause anxiety?} To the best of our knowledge, it does not contain offensive content.

\subsection{Collection process}
\paragraph{How was the data associated with each instance acquired?} The data associated with each instance is directly observable and exported from the raw sensor data (e.g. camera, radar, LiDAR).

\paragraph{What mechanisms or procedures were used to collect the data (e.g., hardware apparatuses or sensors, manual human curation, software programs, software APIs)?} 
\begin{figure}
\begin{floatrow}
\ffigbox{%
  \includegraphics[width=1.13\linewidth]{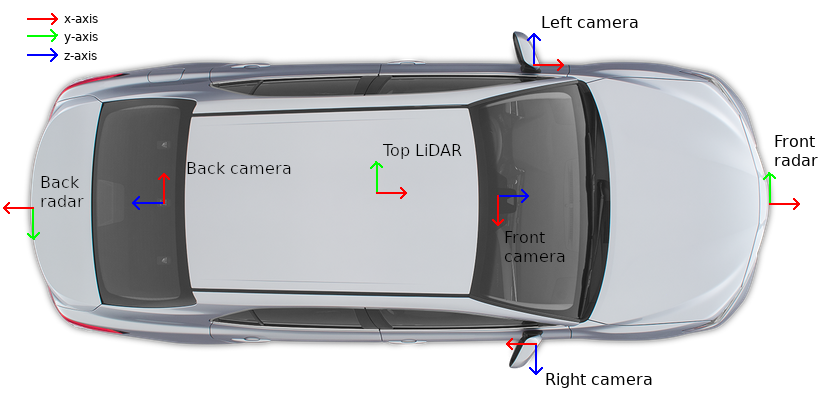}
  
}
{%
  \caption{Sensor setup and coordinate systems.}%
  \label{fig:sensor_setup2}
}

\capbtabbox{%
  \begin{tabular}{c} 
    \includegraphics[width=0.823\linewidth]{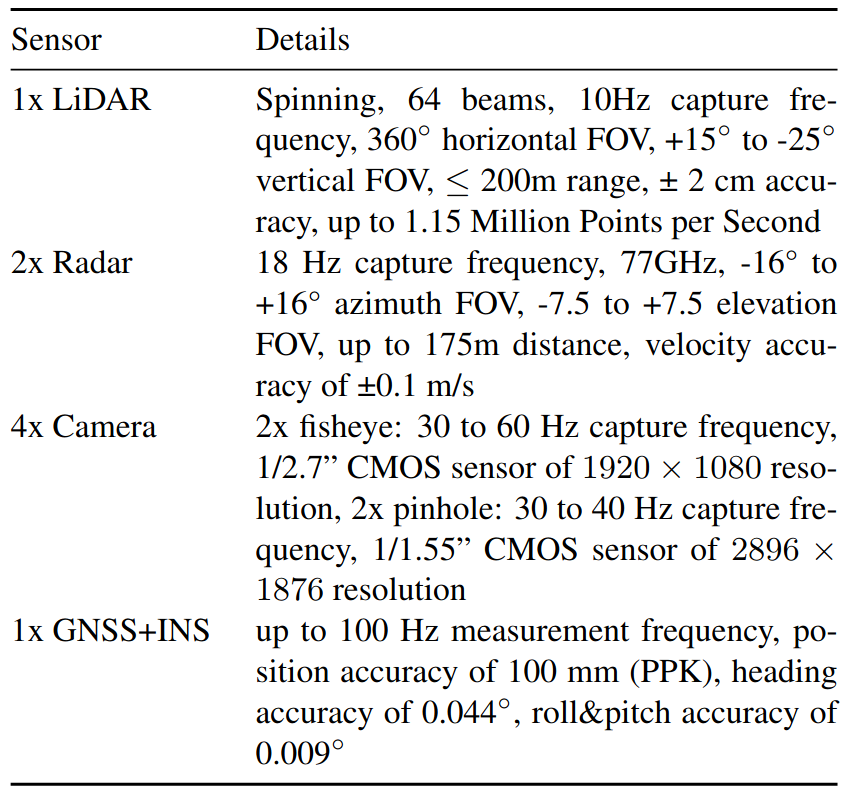} \\
  \end{tabular}
}{%
  \caption{Description of used sensors.}%
  \label{tab:sensor_desc2}
}
\end{floatrow}
\end{figure}

The data was recorded using a roof-mounted, rotating 64-beam LiDAR, four cameras, and two long-range radars, providing 360$^{\circ}$ coverage with sensor redundancy. The localization was based on a high-precision GNSS+INS sensor. Additional details can be found in Figure~\ref{fig:sensor_setup2} and Table~\ref{tab:sensor_desc2}. The data was collected in three countries on two continents with four cars to provide a diverse dataset. The recordings have taken place in California, US; Austria; and Hungary using three Toyota Camry and one Toyota Prius. The validity of sensor data was manually checked using \href{https://github.com/aimotive/aimotive-dataset-loader}{aiMotive Dataset Loader}.

\paragraph{If the dataset is a sample from a larger set, what was the sampling strategy (e.g., deterministic, probabilistic with specific sampling probabilities)?} The dataset has not been sampled from a larger set.

\paragraph{Who was involved in the data collection process (e.g., students, crowdworkers, contractors) and how were they compensated (e.g., how much were crowdworkers paid)?} The data collection process was performed by full-time aiMotive employees. The process did not involve students, crowdworkers, or contractors. The data collection process was part of the employee responsibilities of the participants, who were compensated by monthly salaries.

\paragraph{Over what timeframe was the data collected? Does this timeframe match the creation timeframe of the data associated with the instances (e.g., recent crawl of old news articles)?} The recording phase of the footage was spread across a year to gather data on different seasons and weather conditions. The annotation creation did not depend on the time of raw sensor data recording.

\paragraph{Were any ethical review processes conducted (e.g., by an institutional review board)?} The data collection method has satisfied the requirements given by the Institutional Review Board. The details can be read in aiMotive's \href{https://aimotive.com/documents/d/guest/data-protection-policy-ai-development-5-august-2020-_en_hun_final}{Data Protection Policy}.

\subsection{Preprocessing/cleaning/labeling}
\paragraph{Was any preprocessing/cleaning/labeling of the data done (e.g., discretization or bucketing, tokenization, part-of-speech tagging, SIFT feature extraction, removal of instances, processing of missing values)?} The dataset has been anonymized by blurring faces and license plates. Besides this preprocessing step, no additional method has been applied to raw sensor data.

\paragraph{Was the “raw” data saved in addition to the preprocessed/cleaned/labeled data (e.g., to support unanticipated future uses)? If so, please provide a link or other access point to the “raw” data.} No, the raw data has not been saved to meet GDPR requirements.

\paragraph{Is the software that was used to preprocess/clean/label the data available?} Yes, DashcamCleaner (face and license plate blurring software) is available on \href{https://github.com/tfaehse/DashcamCleaner}{GitHub}.

\subsection{Uses}
\paragraph{Has the dataset been used for any tasks already?} The dataset has been downloaded 100+ times by other researchers and students at the time of submission. Statistics can be found on \href{https://www.kaggle.com/datasets/tamasmatuszka/aimotive-multimodal-dataset}{Kaggle}. 

\paragraph{Is there a repository that links to any or all papers or systems that use the dataset?} Yes, the dataset can be found on \href{https://paperswithcode.com/dataset/aimotive-dataset}{Papers With Code}.

\paragraph{What (other) tasks could the dataset be used for?} The dataset can be used for end-to-end long-range multiple object tracking, motion prediction, visual and LiDAR odometry, and contrastive self-supervised representation learning.

\paragraph{Is there anything about the composition of the dataset or the way it was collected and preprocessed/cleaned/labeled that might impact future uses?} To the best of our knowledge, there is no such case.

\paragraph{Are there tasks for which the dataset should not be used?} The dataset must not be used for any military or harmful application.

\subsection{Distribution}
\paragraph{Will the dataset be distributed to third parties outside of the entity (e.g., company, institution, organization) on behalf of which the dataset was created?} The dataset has been uploaded to \href{https://www.kaggle.com/datasets/tamasmatuszka/aimotive-multimodal-dataset}{Kaggle}. In this way, researchers with limited computational capacities can utilize the Kaggle platform for training models on the dataset.

\paragraph{How will the dataset will be distributed (e.g., tarball on website, API, GitHub)?} Does the dataset have a digital object identifier (DOI)?
The dataset is available on \href{https://github.com/aimotive/aimotive_dataset}{GitHub} and \href{https://www.kaggle.com/datasets/tamasmatuszka/aimotive-multimodal-dataset}{Kaggle}. The digital object identifier (DOI) is 10.34740/kaggle/ds/2738461.

\paragraph{When will the dataset be distributed?} The dataset is already available.

\paragraph{Will the dataset be distributed under a copyright or other intellectual property (IP) license, and/or under applicable terms of use (ToU)?} The dataset is available under Creative Commons' Attribution-NonCommercial-ShareAlike 4.0 International (CC BY-NC-SA 4.0) \href{https://creativecommons.org/licenses/by-nc-sa/4.0/}{licence}.

\paragraph{Have any third parties imposed IP-based or other restrictions on the data associated with the instances?} No.

\paragraph{Do any export controls or other regulatory restrictions apply to the dataset or to individual instances?} No.

\subsection{Maintenance}
\paragraph{Who will be supporting/hosting/maintaining the dataset?} The dataset is supported, hosted, and maintained by aiMotive.

\paragraph{How can the owner/curator/manager of the dataset be contacted (e.g., email address)?} The dataset owner can be contacted at \href{mailto:tamas.matuszka@aimotive.com}{tamas.matuszka@aimotive.com}.

\paragraph{Is there an erratum?} Currently, there is no erratum.

\paragraph{Will the dataset be updated (e.g., to correct labeling errors, add new instances, delete instances)?} The dataset will be updated to correct labeling errors in case of found errors. New instance addition might also happen. The updates will be disseminated on GitHub and Kaggle.

\paragraph{If the dataset relates to people, are there applicable limits on the retention of the data associated with the instances (e.g., were the individuals in question told that their data would be retained for
a fixed period of time and then deleted)?} To the best of our knowledge, there is no such limit.

\paragraph{Will older versions of the dataset continue to be supported/hosted/maintained?} The dataset will continuously be supported/maintained on GitHub.

\paragraph{If others want to extend/augment/build on/contribute to the dataset, is there a mechanism for them to do so?} Contributions are welcome. Pull requests can be sent to our GitHub repositories. In order to be merged into the main branch, at least one review and approval from the code owners are needed. In this way, the contribution can be validated. In terms of dataset augmentation/extension, please contact the dataset owner. One of the main goals of the release of the dataset is to facilitate research in multimodal sensor fusion and robust long-range perception systems. Therefore, building on our dataset is encouraged.

\newpage

\appendix

\section{Appendix}

\subsection{Samples and qualitative results}

\begin{figure}[!h]
  \centering
  \includegraphics[width=1.0\linewidth]{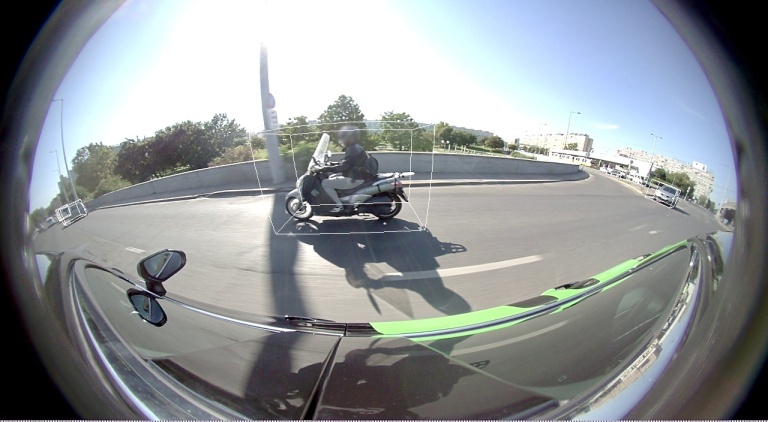}
   \caption{Example motorbike GT created by automatic annotation.}
   \label{fig:ann}
\end{figure}

\begin{figure}
  \centering
   \includegraphics[width=0.95\linewidth]{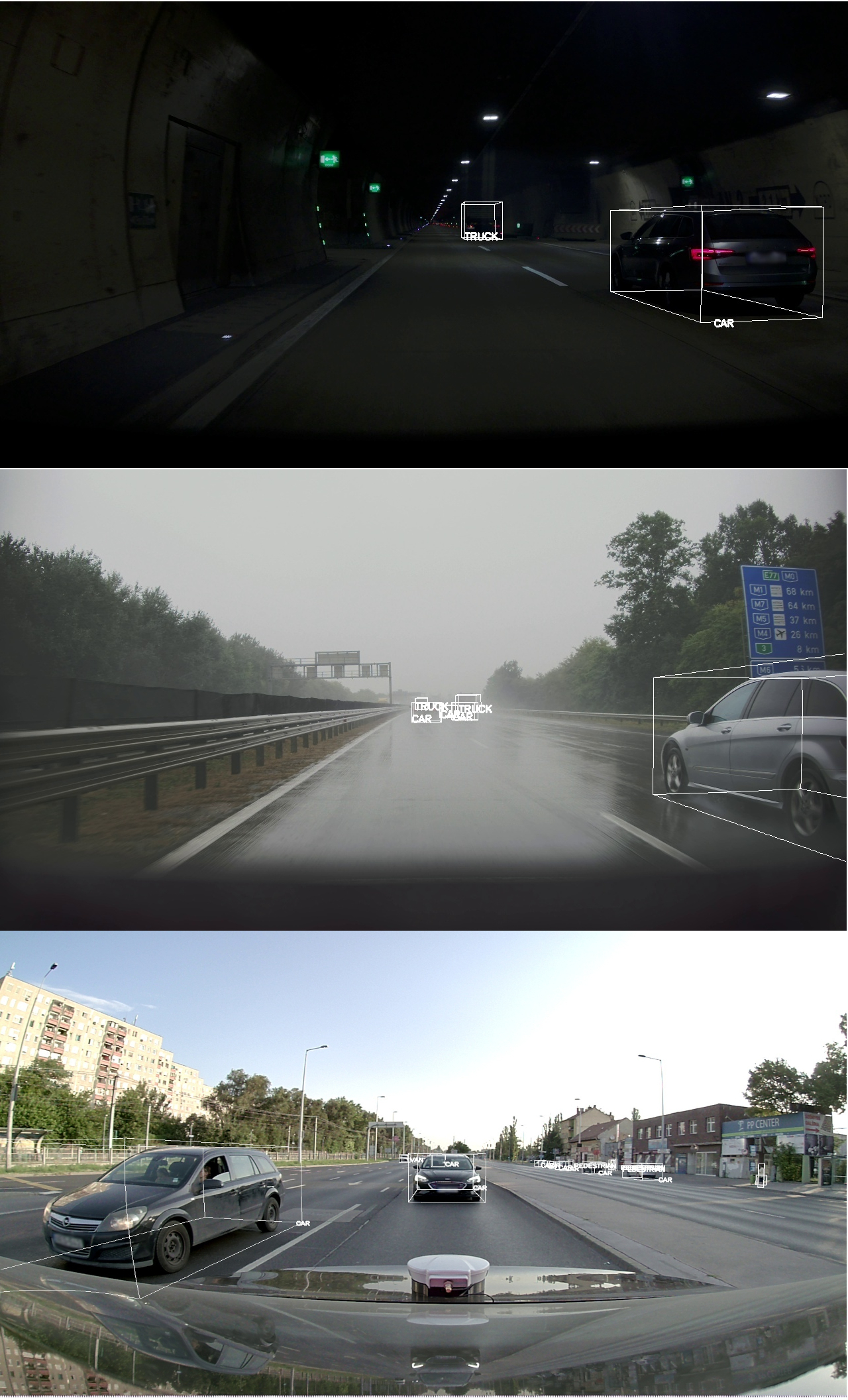}

   \caption{Example ground truth annotations from the training set.}
   \label{fig:annots2}
\end{figure}

\begin{figure}[H]
  \centering
    \includegraphics[width=1.0\linewidth]{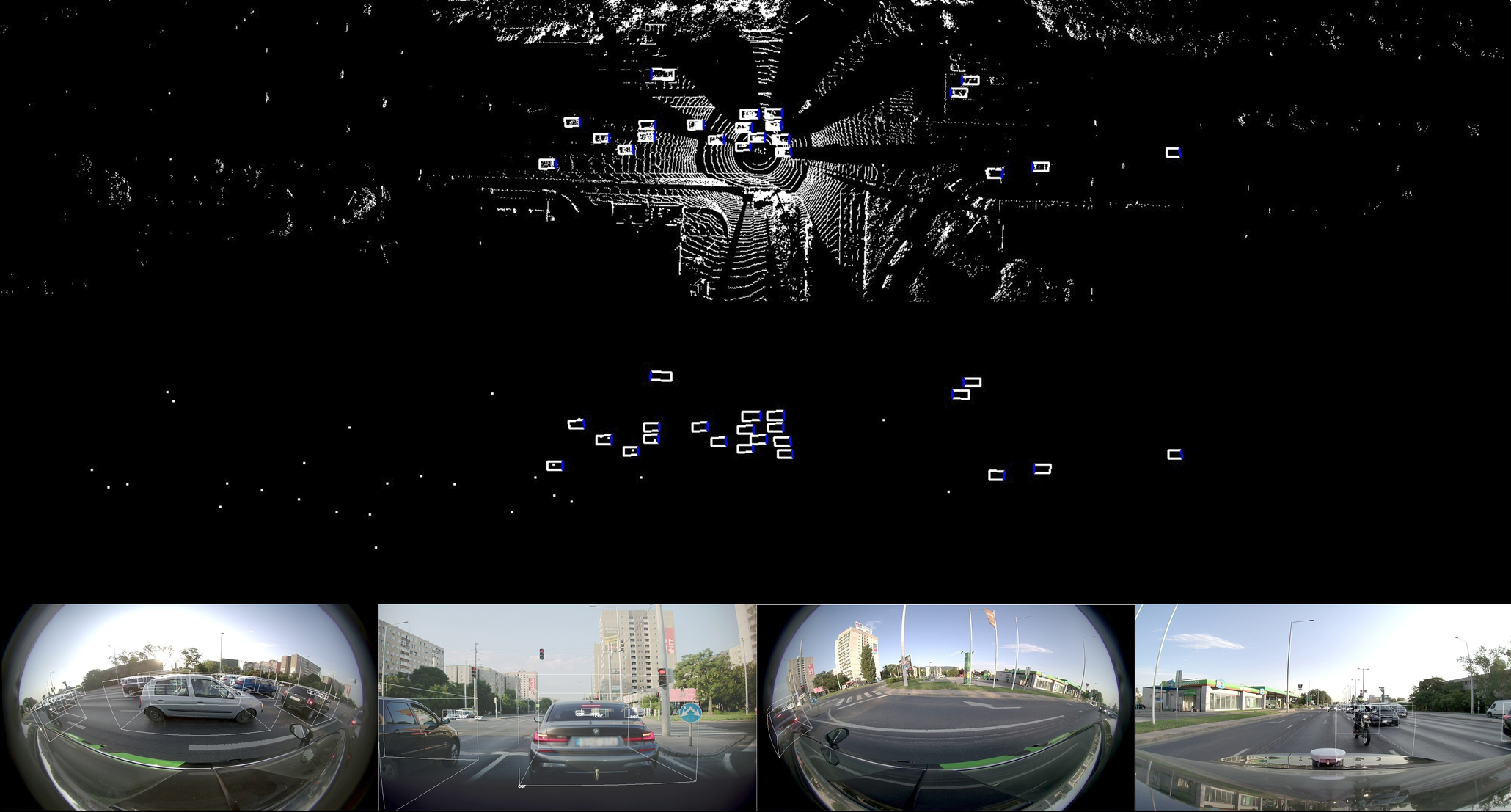}

   \caption{Qualitative results: detections of the LiDAR+radar baseline model on validation set. Top row: detections on LiDAR point cloud. Middle row: detections on radar targets, bottom row (from left to right): detections on left, front, right, and back cameras.}
   \label{fig:detection}
\end{figure}

In this section, we provide qualitative results and several visual examples of the annotations contained by the dataset. Figure~\ref{fig:ann} (VRU annotation) and ~\ref{fig:annots2} (tunnel, heavy rain, dense urban environment) show example annotations generated by our automatic annotation method and manual annotation, respectively. A short video about the camera projection of the annotations created by the automatic annotation method can be seen in this \href{https://adasworks-my.sharepoint.com/:v:/g/personal/tamas_matuszka_aimotive_com1/EY8VeetDvL5PjZx6t7DUQlkBx0Dihh5-Ww5CCmzhiCz61w?e=3O43Ex}{link}. Figure~\ref{fig:detection} visualizes a sample detection of the LiDAR+radar baseline model on the validation set. Note that this sample is not included in the training set. It is the output of one of the baseline models described in the paper.

\begin{figure}[H]
  \centering
    \includegraphics[width=1.0\linewidth]{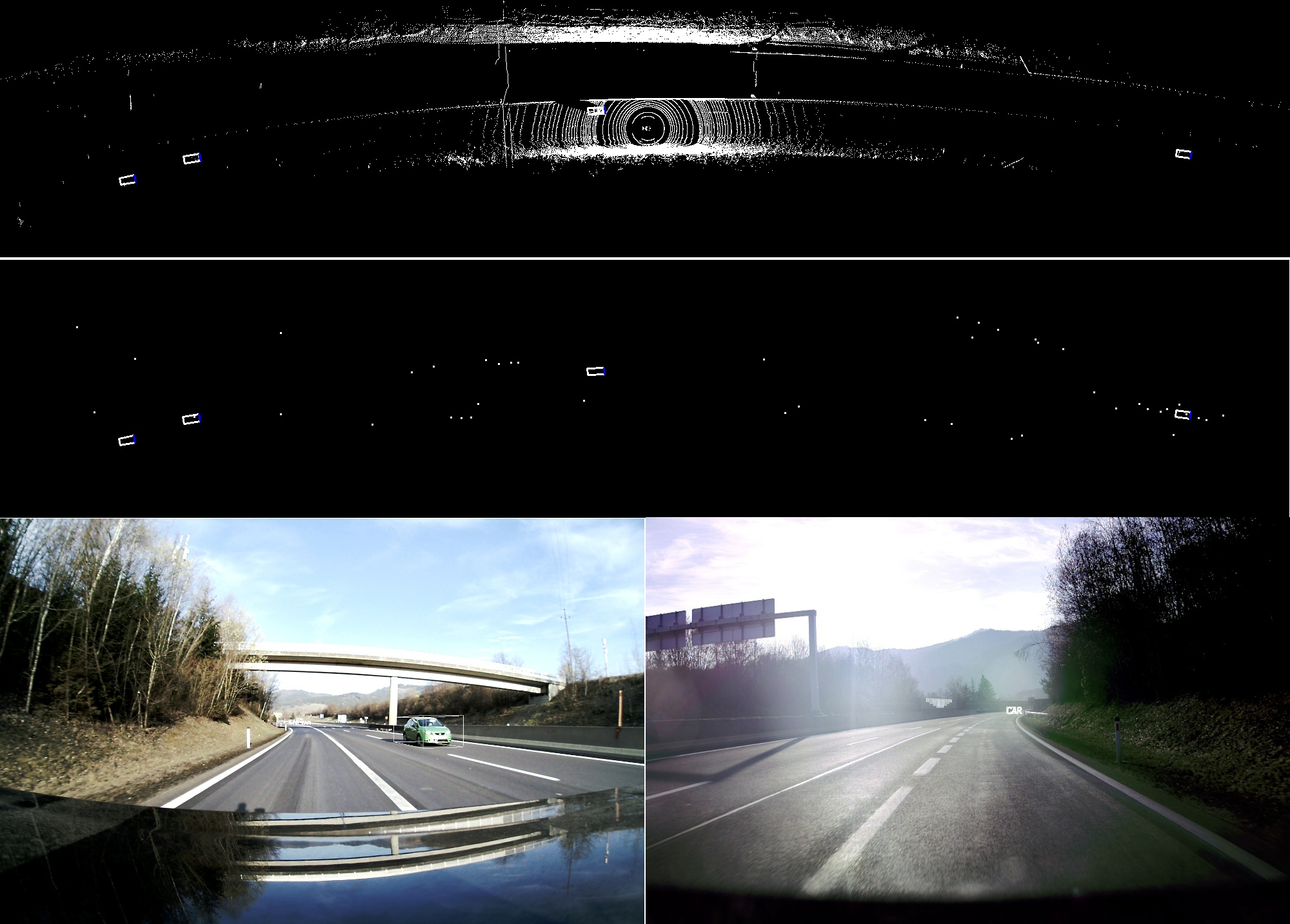}

   \caption{Bounding boxes fitting to individual sensors (best viewed by zooming in) - distant annotations.}
   \label{fig:boxfit}
\end{figure}

\begin{figure}[H]
  \centering
    \includegraphics[width=1.0\linewidth]{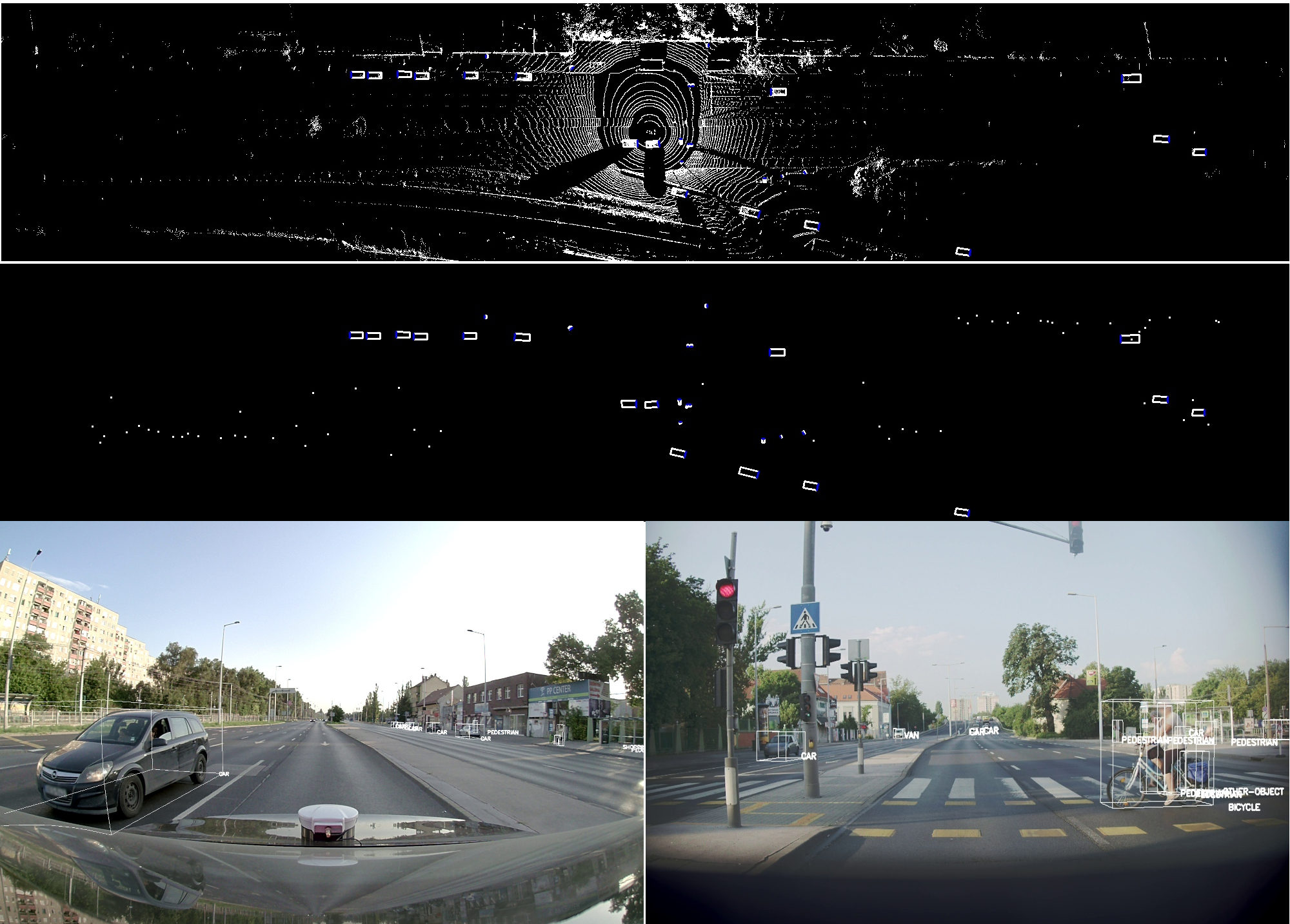}

   \caption{Bounding boxes fitting to individual sensors (best viewed by zooming in) - dense annotations.}
   \label{fig:boxfit2}
\end{figure}

Figure~\ref{fig:boxfit} and ~\ref{fig:boxfit2} visualize how well the bounding boxes fit individual sensor modalities (sensor and bounding box visualization from top to down: LiDAR, radar, camera: back and front). The sensor setup can results in a synchronization limitation caused by the fact that sensors differ in the temporal recording method. Since the rotating LiDAR and rolling shutter cameras have different measurement methods, there are some discrepancies in temporally discretized annotations. This phenomenon is mostly visible when the relative speed difference between the ego and exo car is large. As the figures show, the radar sensor has a visible variance (see distant objects in front of the ego car). The bounding boxes in the distant area might be slightly shifted from the LiDAR point cloud too, as can be seen in the case of the last vehicle behind the ego car in Figure~\ref{fig:boxfit} (best viewed by zooming in).

\subsection{Additional evaluation results}
In this section, we provide additional evaluations of the trained baseline models using the AP-R40, classification accuracy, and Average Orientation Similarity (AOS) metrics. The models were trained for 9 epochs. Furthermore, to investigate the effect of distance on the detection ability, we evaluated our models on various depth ranges (distance range is measured from the ego vehicle, both behind and in front of it).

\begin{table*}[!h]
  \small
  \centering
  \begin{tabular}{@{}l|ccccccc@{}}
    \toprule
    Modalities          & Highway   & Urban     & Night     & Rain      & Full      & Class. Acc.   & AOS\\
    \midrule
    LiDAR               & \bf{0.753}& 0.619     & 0.752     & \bf{0.551}& 0.654     & \bf{0.840}    & 0.731 \\
    LiDAR, camera       & 0.746     & 0.631     & 0.767     & 0.442     & \bf{0.656}& 0.839         & 0.748 \\
    LiDAR, radar        & 0.743     & \bf{0.632}& \bf{0.768}& 0.496     & 0.654     & 0.836         & \bf{0.904} \\
    LiDAR, radar, camera & 0.719    & 0.618     & 0.736     & 0.421     & 0.649     & 0.839         & 0.889 \\
    \bottomrule
  \end{tabular}
  \caption{Comparison of baseline models using AP-R40 (on splits and over the full dataset), classification accuracy, and AOS metrics averaged over val set.}
  \label{tab:metrics_whole}
\end{table*}

\begin{table*}[!h]
  \small
  \centering
  \begin{tabular}{@{}l|ccccccc@{}}
    \toprule
    Modalities          & Highway   & Urban     & Night     & Rain      & Full      & Class. Acc.   & AOS\\
    \midrule
    LiDAR               & \bf{0.497}& 0.446     & 0.437     & 0.587     & 0.459     & \bf{0.756}    & 0.858 \\
    LiDAR, camera       & 0.494     & 0.452     & 0.499     & 0.495     & 0.462     & 0.751         & 0.793 \\
    LiDAR, radar        & 0.477     & \bf{0.490}& \bf{0.512}& \bf{0.630}& \bf{0.495}& 0.749         & 0.869 \\
    LiDAR, radar, camera & 0.461    & 0.461     & 0.474     & 0.410     & 0.464     & 0.734         & \bf{0.876} \\
    \bottomrule
  \end{tabular}
  \caption{Comparison of baseline models in the distant region (\textgreater75m) using AP-R40 (on splits and over the full dataset), classification accuracy, and AOS metrics averaged over val set.}
  \label{tab:metrics_distant}
\end{table*}

\newpage
\subsubsection{Range-based evaluation: [0-30) m}

\begin{table*}[!h]
  \small
  \centering
  \begin{tabular}{@{}l|cccc@{}}
    \toprule
    Modalities          & Highway   & Urban     & Night     & Rain  \\
    \midrule
    LiDAR               & \bf{0.937}& 0.617     & 0.844     & \bf{0.313} \\
    LiDAR, camera       & 0.911     & \bf{0.652}& \bf{0.847}& 0.292 \\
    LiDAR, radar        & 0.932     & 0.631     & 0.824     & 0.195 \\
    LiDAR, radar, camera & 0.906    & 0.621     & 0.802     & 0.233 \\
    \bottomrule
  \end{tabular}
  \caption{Comparison of baseline models in the range of [0-30) m from ego car using all points interpolation AP metric.}
  \label{tab:metrics0}
\end{table*}

\begin{table*}[!h]
  \small
  \centering
  \begin{tabular}{@{}l|cccc@{}}
    \toprule
    Modalities          & Highway   & Urban     & Night     & Rain \\
    \midrule
    LiDAR               & \bf{0.894}& 0.605     & 0.823     & \bf{0.325} \\
    LiDAR, camera       & 0.876     & \bf{0.620}& \bf{0.828}& \bf{0.325} \\
    LiDAR, radar        & 0.890     & 0.615     & 0.773     & 0.236 \\
    LiDAR, radar, camera & 0.877    & 0.608     & 0.772     & 0.273 \\
    \bottomrule
  \end{tabular}
  \caption{Comparison of baseline models in the range of [0-30) m from ego car using 11-points AP metric.}
  \label{tab:metrics2}
\end{table*}

\begin{table*}[!h]
  \small
  \centering
  \begin{tabular}{@{}l|cccccc@{}}
    \toprule
    Modalities          & Highway   & Urban     & Night     & Rain \\
    \midrule
    LiDAR               & \bf{0.925}& 0.606     & 0.829     & \bf{0.302} \\
    LiDAR, camera       & 0.901     & \bf{0.639}& \bf{0.833}& 0.284 \\
    LiDAR, radar        & 0.921     & 0.616     & 0.809     & 0.187 \\
    LiDAR, radar, camera & 0.891    & 0.610     & 0.791     & 0.225 \\
    \bottomrule
  \end{tabular}
  \caption{Comparison of baseline models in the range of [0-30) m from ego car using AP-R40 metric.}
  \label{tab:metrics3}
\end{table*}

\begin{table*}[!h]
  \small
  \centering
  \begin{tabular}{@{}l|ccccc@{}}
    \toprule
    Modalities          & AP-all    & AP-11     & AP-R40    & Class. Acc.   & AOS\\
    \midrule
    LiDAR               & 0.671     & 0.665     & 0.654     & 0.863         & 0.937 \\
    LiDAR, camera       & \bf{0.691}& \bf{0.681}& \bf{0.677}& 0.863         & 0.866 \\
    LiDAR, radar        & 0.674     & 0.669     & 0.659     & 0.858         & \bf{0.913} \\
    LiDAR, radar, camera & 0.664    & 0.662     & 0.652     & \bf{0.871}    & 0.867 \\
    \bottomrule
  \end{tabular}
  \caption{Comparison of baseline models in the range of [0-30) m from ego car on the whole dataset using all points, 11-points AP, and AP-R40, classification accuracy, and AOS metrics.}
  \label{tab:metrics4}
\end{table*}

\newpage
%%%%%%%%%%%%%%%%%%%%%%%%%%%%%%%% 30 - 60 %%%%%%%%%%%%%%%%%%%%%%%%%%%%%%%%%
\subsubsection{Range-based evaluation: [30-60) m}

\begin{table*}[!h]
  \small
  \centering
  \begin{tabular}{@{}l|cccc@{}}
    \toprule
    Modalities          & Highway   & Urban     & Night     & Rain  \\
    \midrule
    LiDAR               & \bf{0.933}& 0.672     & 0.766     & \bf{0.772} \\
    LiDAR, camera       & 0.922     & \bf{0.676}& \bf{0.796}& 0.632 \\
    LiDAR, radar        & 0.924     & 0.664     & 0.790     & 0.686 \\
    LiDAR, radar, camera & 0.924    & 0.670     & 0.786     & 0.645 \\
    \bottomrule
  \end{tabular}
  \caption{Comparison of baseline models in the range of [30-60) m from ego car using all points interpolation AP metric.}
  \label{tab:metrics5}
\end{table*}

\begin{table*}[!h]
  \small
  \centering
  \begin{tabular}{@{}l|cccc@{}}
    \toprule
    Modalities          & Highway   & Urban     & Night     & Rain \\
    \midrule
    LiDAR               & \bf{0.888}& 0.664     & 0.739     & \bf{0.751} \\
    LiDAR, camera       & 0.879     & \bf{0.668}& \bf{0.769}& 0.621 \\
    LiDAR, radar        & 0.886     & 0.625     & 0.760     & 0.680 \\
    LiDAR, radar, camera & 0.883    & 0.663     & \bf{0.769}& 0.623 \\
    \bottomrule
  \end{tabular}
  \caption{Comparison of baseline models in the range of [30-60) m from ego car using 11-points AP metric.}
  \label{tab:metrics6}
\end{table*}

\begin{table*}[!h]
  \small
  \centering
  \begin{tabular}{@{}l|cccccc@{}}
    \toprule
    Modalities          & Highway   & Urban     & Night     & Rain \\
    \midrule
    LiDAR               & \bf{0.920}& 0.661     & 0.754     & \bf{0.761} \\
    LiDAR, camera       & 0.910     & \bf{0.662}& \bf{0.784}& 0.621 \\
    LiDAR, radar        & 0.913     & 0.649     & 0.777     & 0.675 \\
    LiDAR, radar, camera & 0.914    & \bf{0.662}& 0.776     & 0.633 \\
    \bottomrule
  \end{tabular}
  \caption{Comparison of baseline models in the range of [30-60) m from ego car using AP-R40 metric.}
  \label{tab:metrics7}
\end{table*}

\begin{table*}[!h]
  \small
  \centering
  \begin{tabular}{@{}l|ccccc@{}}
    \toprule
    Modalities          & AP-all    & AP-11     & AP-R40    & Class. Acc.   & AOS\\
    \midrule
    LiDAR               & 0.722     & 0.697     & 0.706     & 0.873         & \bf{0.944} \\
    LiDAR, camera       & \bf{0.723}& 0.695     & \bf{0.712}& 0.873         & 0.892 \\
    LiDAR, radar        & 0.716     & 0.696     & 0.704     & 0.873         & 0.924 \\
    LiDAR, radar, camera & 0.722    & \bf{0.702}& 0.711     & 0.873         & 0.931 \\
    \bottomrule
  \end{tabular}
  \caption{Comparison of baseline models in the range of [30-60) m from ego car on the whole dataset using all points, 11-points AP, and AP-R40, classification accuracy, and AOS metrics.}
  \label{tab:metrics8}
\end{table*}

\newpage
%%%%%%%%%%%%%%%%%%%%%%%%%%%%%%%% 60 - 120 %%%%%%%%%%%%%%%%%%%%%%%%%%%%%%%%%
\subsubsection{Range-based evaluation: (60-120] m}
\begin{table*}[!h]
  \small
  \centering
  \begin{tabular}{@{}l|cccc@{}}
    \toprule
    Modalities          & Highway   & Urban     & Night     & Rain  \\
    \midrule
    LiDAR               & 0.773     & 0.612     & 0.667     & \bf{0.755} \\
    LiDAR, camera       & 0.796     & 0.615     & 0.690     & 0.700 \\
    LiDAR, radar        & 0.776     & 0.629     & \bf{0.703}& 0.739 \\
    LiDAR, radar, camera &\bf{0.813}&\bf{0.637} & 0.669     & 0.594 \\
    \bottomrule
  \end{tabular}
  \caption{Comparison of baseline models in the range of [60-120) m from ego car using all points interpolation AP metric.}
  \label{tab:metrics_60_120}
\end{table*}

\begin{table*}[!h]
  \small
  \centering
  \begin{tabular}{@{}l|cccc@{}}
    \toprule
    Modalities          & Highway   & Urban     & Night     & Rain \\
    \midrule
    LiDAR               & 0.739     & 0.586     & 0.656     & \bf{0.727} \\
    LiDAR, camera       & 0.772     & 0.599     & 0.680     & 0.684 \\
    LiDAR, radar        & 0.740     & \bf{0.613}& \bf{0.694}& 0.723 \\
    LiDAR, radar, camera &\bf{0.796}& \bf{0.613}& 0.646     & 0.575 \\
    \bottomrule
  \end{tabular}
  \caption{Comparison of baseline models in the range of [60-120) m from ego car using 11-points AP metric.}
  \label{tab:metrics9}
\end{table*}

\begin{table*}[!h]
  \small
  \centering
  \begin{tabular}{@{}l|cccccc@{}}
    \toprule
    Modalities          & Highway   & Urban     & Night     & Rain \\
    \midrule
    LiDAR               & 0.761     & 0.596     & 0.657     & \bf{0.743} \\
    LiDAR, camera       & 0.786     & 0.599     & 0.678     & 0.690 \\
    LiDAR, radar        & 0.764     & 0.615     & \bf{0.692}& 0.730 \\
    LiDAR, radar, camera &\bf{0.804}&\bf{0.625} & 0.654     & 0.578 \\
    \bottomrule
  \end{tabular}
  \caption{Comparison of baseline models in the range of [60-120) m from ego car using AP-R40 metric.}
  \label{tab:metrics10}
\end{table*}

\begin{table*}[!h]
  \small
  \centering
  \begin{tabular}{@{}l|ccccc@{}}
    \toprule
    Modalities          & AP-all    & AP-11     & AP-R40    & Class. Acc.   & AOS\\
    \midrule
    LiDAR               & 0.639     & 0.637     & 0.624     & \bf{0.798}    & 0.834 \\
    LiDAR, camera       & 0.631     & 0.629     & 0.618     & \bf{0.798}    & 0.803 \\
    LiDAR, radar        & 0.659     & 0.628     & 0.642     & 0.793         & 0.871 \\
    LiDAR, radar, camera &\bf{0.668}& \bf{0.665}& \bf{0.656}& 0.790         & \bf{0.874} \\
    \bottomrule
  \end{tabular}
  \caption{Comparison of baseline models in the range of [60-120) m from ego car on the whole dataset using all points, 11-points AP, and AP-R40, classification accuracy, and AOS metrics.}
  \label{tab:metrics_60_120_all}
\end{table*}

\newpage
%%%%%%%%%%%%%%%%%%%%%%%%%%%%%%%% 120 > %%%%%%%%%%%%%%%%%%%%%%%%%%%%%%%%%
\subsubsection{Range-based evaluation: [120-200] m}
\begin{table*}[!h]
  \small
  \centering
  \begin{tabular}{@{}l|cccc@{}}
    \toprule
    Modalities          & Highway   & Urban     & Night     & Rain  \\
    \midrule
    LiDAR               & \bf{0.307}& 0.194     & 0.132     & 0.316 \\
    LiDAR, camera       & 0.295     & 0.184     & 0.183     & 0.154 \\
    LiDAR, radar        & 0.289     & \bf{0.272}& 0.253     & \bf{0.402} \\
    LiDAR, radar, camera & 0.205    & 0.157     & \bf{0.277}& 0.155 \\
    \bottomrule
  \end{tabular}
  \caption{Comparison of baseline models in the range of [120-200] m from ego car using all points interpolation AP metric.}
  \label{tab:metrics11}
\end{table*}

\begin{table*}[!h]
  \small
  \centering
  \begin{tabular}{@{}l|cccc@{}}
    \toprule
    Modalities          & Highway   & Urban     & Night     & Rain \\
    \midrule
    LiDAR               & \bf{0.341}& 0.217     & 0.133     & 0.334 \\
    LiDAR, camera       & 0.316     & 0.220     & 0.203     & 0.199 \\
    LiDAR, radar        & 0.310     & \bf{0.260}& 0.270     & \bf{0.395} \\
    LiDAR, radar, camera & 0.239    & 0.186     & \bf{0.292}& 0.167 \\
    \bottomrule
  \end{tabular}
  \caption{Comparison of baseline models in the range of [120-200] m from ego car using 11-points AP metric.}
  \label{tab:metrics12}
\end{table*}

\begin{table*}[!h]
  \small
  \centering
  \begin{tabular}{@{}l|cccccc@{}}
    \toprule
    Modalities          & Highway   & Urban     & Night     & Rain \\
    \midrule
    LiDAR               & \bf{0.301}& 0.187     & 0.128     & 0.302 \\
    LiDAR, camera       & 0.286     & 0.170     & 0.176     & 0.144 \\
    LiDAR, radar        & 0.280     & \bf{0.262}& 0.242     & \bf{0.389} \\
    LiDAR, radar, camera & 0.187    & 0.148     & \bf{0.265}& 0.149 \\
    \bottomrule
  \end{tabular}
  \caption{Comparison of baseline models in the range of [120-200] m from ego car using AP-R40 metric.}
  \label{tab:metrics13}
\end{table*}

\begin{table*}[!h]
  \small
  \centering
  \begin{tabular}{@{}l|ccccc@{}}
    \toprule
    Modalities          & AP-all    & AP-11     & AP-R40    & Class. Acc.   & AOS\\
    \midrule
    LiDAR               & 0.249     & 0.272     & 0.237     & \bf{0.750}    & 0.883 \\
    LiDAR, camera       & 0.235     & 0.264     & 0.224     & 0.726         & 0.784 \\
    LiDAR, radar        & \bf{0.279}& \bf{0.296}& \bf{0.271}& 0.729         & \bf{0.901} \\
    LiDAR, radar, camera & 0.187    & 0.222     & 0.178     & 0.673         & 0.884 \\
    \bottomrule
  \end{tabular}
  \caption{Comparison of baseline models in the range of [120-200] m from ego car on the whole dataset using all points, 11-points AP, and AP-R40, classification accuracy, and AOS metrics.}
  \label{tab:metrics14}
\end{table*}

%%%%%%%%%%%%%%%%%%%%%%%%%%%%%%%% 120 > %%%%%%%%%%%%%%%%%%%%%%%%%%%%%%%%%
\subsubsection{Camera-only baseline model}
The camera-only baseline model is adapted from BEVDepth repository\footnote{\href{https://github.com/Megvii-BaseDetection/BEVDepth}{https://github.com/Megvii-BaseDetection/BEVDepth}}. We changed the image backbone from ResNet-50 to ResNet-152 to utilize a stronger image encoder. This baseline model performed very poorly (<0.10 AP-R40) on the whole perception range ($\pm200$ meters in longitudinal, $\pm25.6$ meters in lateral direction). We investigated the heatmap tensors and found that the model is very uncertain in distant regions. This fact results in a large number of false positive detections. This might explain the unexpected multimodal results in some cases since this phenomenon might hinder the multimodal feature fusion. The AP-R40 performance in the near-region (<30 m) is also worse than our other unimodal baseline model, though (camera-only: 0.406,  LiDAR-only: 0.677).

Nevertheless, the combination of the camera-only model with additional sensor modalities resulted in better performance on the urban/night splits and the whole dataset (on the entire perception range) using the AP-R40 metrics (Table~\ref{tab:metrics_whole}). The improvement is even more visible in distant regions (Table~\ref{tab:metrics_distant}, Table~\ref{tab:metrics_60_120}-\ref{tab:metrics_60_120_all}). Finally, we would like to solicit contributions from the research community in order to develop more performant unimodal and multimodal models.

\end{document}